\documentclass[12pt]{article}

\usepackage[margin=1in]{geometry}
\usepackage{setspace}
\usepackage{graphicx}
\usepackage{booktabs}
\usepackage{amsmath}
\usepackage{amssymb}
\usepackage{natbib}
\usepackage{hyperref}
\usepackage{xcolor}
\usepackage{lineno}
\usepackage{caption}
\usepackage{subcaption}
\usepackage{float}

\hypersetup{
    colorlinks=true,
    linkcolor=black,
    citecolor=black,
    urlcolor=blue
}

\setcitestyle{numbers,square}

\title{\textbf{Elite political incivility is rising across democracies}}

\author{
Petter Törnberg$^{1,*}$ \and Juliana Chueri$^{2}$
}

\date{ \small
$^1$ ILLC, University of Amsterdam. 
$^2$ Department of Political Science and Public Administration, Vrije Universiteit Amsterdam. 
$^*$ Correspondence: p.tornberg@uva.nl
}

\begin{document}
\maketitle

\begin{abstract}
\noindent Is political incivility rising across democracies---and if so, why? Analysing approximately 13.8 million tweets from parliamentarians in 26 countries with a validated large language model, we find that the share of severe incivility nearly doubled between 2017 and 2022, from $\approx$2.7\% to $\approx$5.8\% of tweets. Populism is the strongest party-level predictor (OR~=~1.46, 95\%~CI: 1.29--1.65), and its association with incivility is amplified in opposition. Stronger liberal democracy is associated with lower severe incivility (OR~=~0.84 per SD, 95\%~CI: 0.72--0.98). The rise in incivility is overwhelmingly \textit{behavioral}---party families across the spectrum becoming more aggressive---rather than \textit{compositional}: the growing prominence of radical parties explains little of it. Nationalist and radical-right parties remain the most uncivil throughout, but the gap narrows as their 2020 peak partially recedes while mainstream parties become steadily more uncivil. Together, these findings point to a \textit{mainstreaming of incivility}: aggressive and demeaning rhetoric is becoming increasingly common across the political spectrum, with implications for public trust and democratic resilience.
\end{abstract}

\clearpage

\section{Introduction}\label{sec:intro}
Heated exchanges have always been part of democratic politics. But political incivility---hostile, disrespectful, or contemptuous language directed at opponents, institutions, or social groups~\citep{mutz2005new,theocharis2020dynamics}---may be becoming a more routine feature of political competition. Scholars, journalists, and citizens have expressed alarm at an apparent erosion of civil political discourse~\citep{haidt2012look,iyengar2012affect,theocharis2020dynamics,mason2018uncivil}. The stakes of that erosion are well documented: reduced trust in political institutions~\citep{mutz2005new}, psychological harm to those targeted~\citep{kasturiratna2025umbrella}, intensified partisan polarization~\citep{skytte2021dimensions}, and greater tolerance for anti-democratic behavior~\citep{gervais2019rousing}. Incivility among political elites is especially consequential because politicians help set the norms of public discourse more broadly: when they use aggressive and demeaning language, they signal that such behavior is an acceptable part of political life.


Yet despite widespread concern, the comparative and longitudinal evidence behind this narrative remains thin. Has political incivility actually increased? If so, does this reflect uncivil parties gaining prominence, or parties themselves becoming more aggressive? Is any such rise concentrated at the populist and radical-right margins of the political spectrum, or is it spreading across the mainstream? And is elite incivility linked to the health of democratic institutions?

Existing research leaves these questions open for three reasons. Most studies focus on a single country, typically the United States, limiting generalizability and precluding systematic cross-national comparison~\citep{frimer2023incivility}. Most are also cross-sectional or cover short time windows, making it difficult to establish whether incivility is genuinely rising rather than simply fluctuating~\citep{theocharis2020dynamics,klinger2023campaigns}. And the automated measures typically used are trained on English-language web content, raising doubts about their validity for elite political discourse across languages~\citep{nogara2025toxic,hartmann2026bye}.

We address all three limitations at once. Analysing approximately 13.8 million original tweets from parliamentarians in 26 democracies between May 2017 and March 2022, labelled for incivility using a multilingual large language model (LLM) validated against human coding, we fit a hierarchical Bayesian beta-binomial model that accounts for variation across parties, countries, and time. This design lets us estimate the magnitude of any trend, identify its party- and country-level correlates, and---critically---distinguish behavioral change from shifts resulting from the composition of the party system.

Several theoretical expectations organize the analysis. Incivility carries real costs, including reputational damage and erosion of institutional trust~\citep{mutz2005new,frimer2018montagu}, so its persistence and apparent rise imply that its benefits outweigh those costs in at least some contexts. One source of benefit is structural: where platform algorithms reward engagement, inflammatory content earns disproportionate reach~\citep{brady2017emotion,tornberg2022digital}, creating a standing incentive to be uncivil regardless of party. A second source is strategic and varies by position: opposition parties have more to gain from disrupting incumbents and less to lose from breaking norms, while governing parties face moderating pressure from the responsibilities of office---the \textit{inclusion-moderation thesis}~\citep{akkerman2016radical}. Populist parties may face a still sharper version of this incentive, since confrontational language is not incidental to their appeal but aligns directly with their core framing of politics as a struggle between a virtuous people and a corrupt elite~\citep{engesser2017populism,tornberg2025lie}.

A further distinction concerns \textit{how} any aggregate rise comes about, and the two possibilities carry very different implications. Incivility may rise \textit{compositionally}, simply because already-uncivil parties gain electoral prominence, or \textit{behaviorally}, because parties themselves become more aggressive over time. If \textit{rhetorical accommodation} is underway---mainstream parties adopting a more confrontational style in response to radical competitors~\citep{klinger2023campaigns}---the rise should show up as behavioral change concentrated in the mainstream, narrowing but not eliminating the gap with the radical right. Finally, we ask whether democratic quality is associated with elite incivility at all, treating the two as potentially mutually reinforcing while remaining agnostic about the direction of causality~\citep{levitsky2018how,dodeigne2025hostility}.

Four main findings emerge. Political incivility nearly doubled over the five-year period, from approximately 2.7\% to 5.8\% of parliamentarian tweets. This rise is overwhelmingly \textit{behavioral} rather than \textit{compositional}: party families across the spectrum became more uncivil over time, while the growing prominence of already-uncivil parties explains little of the increase. Mainstream families---including the radical left, conservatives, and social democrats---account for much of the rise in absolute terms, narrowing the gap with nationalist and radical-right parties, which remain the most uncivil throughout even as their 2020 peak partially recedes. Populism is the strongest party-level predictor of incivility, and this association is amplified in opposition, consistent with a strategic account of incivility as a tool of political disruption. And weaker liberal democracy is associated with higher severe incivility, linking institutional context directly to the rhetorical norms of political communication. These findings hold across a broad range of alternative specifications: the main structural effects replicate at a lower incivility threshold, are corroborated by an independent measure for the party-level predictors, and survive replacing the GPT-4.1 classifier with an open-source XLM-RoBERTa-large model calibrated against human-coded tweets.

Together, these findings describe a \textit{mainstreaming of incivility}: aggressive and demeaning rhetoric is no longer confined to the populist and radical-right margins of the political spectrum but is becoming increasingly common across mainstream parties as well. This is not convergence to parity: nationalist and radical-right parties remain considerably more uncivil than the rest of the field throughout the period. But the trajectory is unmistakable---the gap is narrowing, not because the radical right is moderating, but because other party families are catching up.

\section{Results}\label{sec:results}

\subsection*{Political incivility increased substantially between 2017 and 2022}

\begin{figure}[htbp]
\centering
\includegraphics[width=0.80\textwidth]{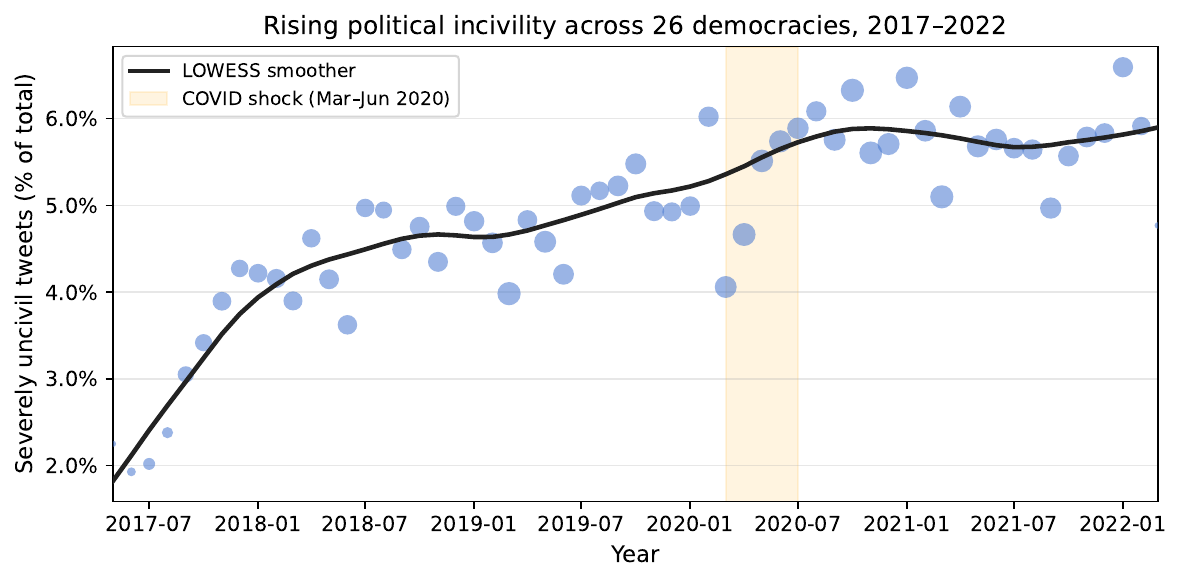}
\caption{\textbf{Rising political incivility across 26 democracies, 2017--2022.} Monthly share of severe incivility (score $\geq 2$), aggregated across all parties and countries and weighted by tweet volume. Points are scaled by total tweet volume; the curve is a LOWESS smoother (bandwidth = 0.25). Comparing the 2017 average with the 2021--2022 average---the convention used throughout because literal smoother endpoints are more edge-sensitive (Supplementary Information Section~3)---incivility more than doubled, from $\approx$2.7\% to $\approx$5.8\%. The shaded band marks the COVID-19 shock period (March--June 2020); the visible dip is consistent with a temporary rally-around-the-flag effect (see text and Supplementary Figure~\ref{fig:si_threshold_comp}).}
\label{fig:timetrend}
\end{figure}

Severe political incivility more than doubled over the study period (Figure~\ref{fig:timetrend}). Its aggregate share rose from approximately 2.7\% of tweets in 2017 to 5.8\% by 2021--2022. This threshold (score $\geq 2$ on a 0--3 scale) captures only the most hostile content: approximately 22\% of tweets contain \textit{any} incivility ($\geq 1$), rising from $\approx$17.9\% to $\approx$23.5\% over the same period. Severe incivility therefore represents roughly the most hostile quarter of all uncivil content (Supplementary Information Section~3).

A visible dip coincides with the onset of the COVID-19 pandemic in early 2020, consistent with ``rally-around-the-flag'' dynamics in which external threats temporarily suppress partisan conflict~\citep{mueller1970presidential,baum2002rallying}. The decline is credible at the any-incivility threshold (OR~=~0.87, 95\%~CI: 0.81--0.93) and directionally similar, though less precisely estimated, for severe incivility (OR~=~0.91, 95\%~CI: 0.82--1.01). One possibility is that the initially unifying effect of the pandemic was partly offset by the subsequent politicization of pandemic management in some countries.

The hierarchical Bayesian model confirms the broader upward trend. The global time slope is OR~=~1.15 per standard deviation of time (95\%~CI: 1.08--1.21), after accounting for party and country differences, cabinet status, election periods, and pandemic-related shocks. Election periods show no credible association with severe incivility (OR~=~0.94, 95\%~CI: 0.87--1.01). The post-COVID period is associated with modestly lower incivility than the pre-pandemic baseline (OR~=~0.91, 95\%~CI: 0.83--0.99), but this leveling occurs well above the 2017 level. The long-run rise therefore remains clear.

\subsection*{The rise is widespread across countries but heterogeneous in magnitude}

\begin{figure}[htbp]
\centering
\includegraphics[width=0.45\textwidth]{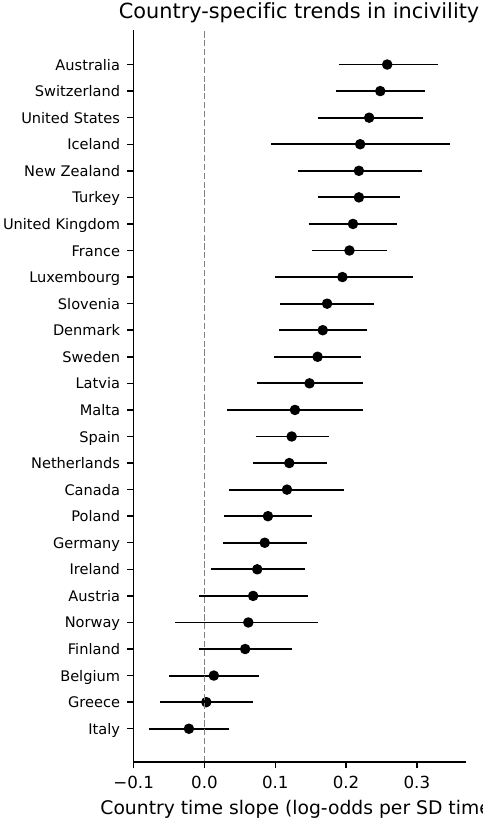}
\caption{\textbf{Country-specific incivility time slopes.} Posterior medians and 95\% credible intervals of country-specific time slopes (global slope $\beta_t$ plus country random slope $b_c$) from the main model, sorted by posterior median. Most countries show positive estimated trends, although the magnitude of change varies substantially and a small number are flat or slightly declining.}
\label{fig:extdata_country}
\end{figure}

The increase extends across much of the sample. Figure~\ref{fig:extdata_country} shows country-specific time slopes from the hierarchical model, capturing change across each country's full monthly trajectory. Most countries have positive estimated slopes, although the magnitude of change varies substantially.

Descriptively, the largest start-to-end increases in the LOWESS-smoothed series occur in Spain ($+$4.7~pp), Australia ($+$4.7~pp), Switzerland ($+$3.7~pp), and Slovenia ($+$3.4~pp). Greece is the only country with a clear decline ($-$4.7~pp), plausibly related to the collapse of Golden Dawn following the 2020 conviction of its leadership and the party's classification as a criminal organization. Austria is essentially flat ($+$0.6~pp). These descriptive rankings need not coincide with the modeled slopes in Figure~\ref{fig:extdata_country}: Italy and Greece show the most notable divergences between the two measures (Supplementary Information Section~1). This heterogeneity is consistent with the cross-national variation in liberal democracy, populist party strength, and specific political events that characterizes our sample (see below). Overall, the rise is widespread but heterogeneous, with positive trends in the large majority of countries.

\subsection*{The rise is driven mainly by behavioral change, not compositional change}

The aggregate rise could reflect two distinct processes: it may be \textit{compositional}, if already-uncivil parties come to account for a larger share of political communication, or \textit{behavioral}, if parties themselves become more uncivil. The latter would indicate a broader shift in the rhetorical norms of political competition. To separate the two processes, we apply a shift-share decomposition of the annual trend (see Methods).

\begin{figure}[htbp]
\centering
\includegraphics[width=0.85\textwidth]{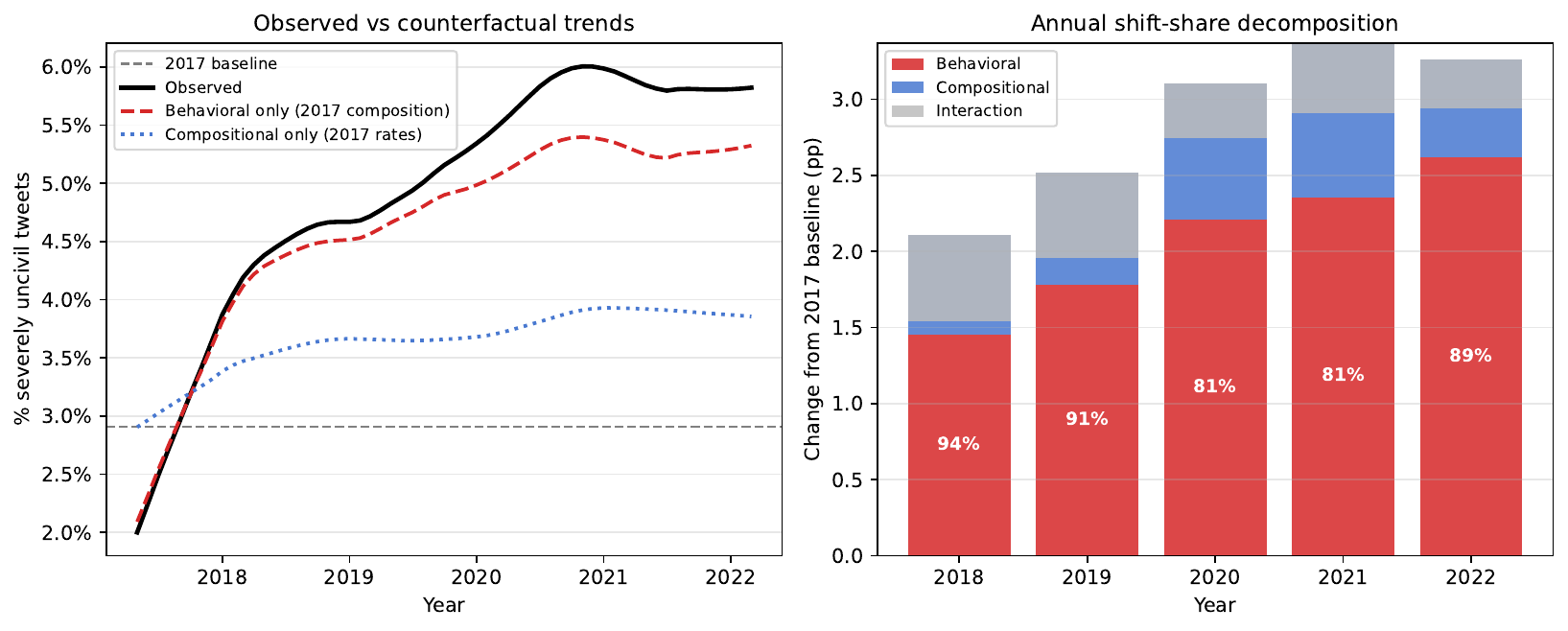}
\caption{\textbf{Shift-share decomposition of the incivility trend.} \textit{Left}: Observed monthly incivility rate (black) compared with two counterfactuals---behavioral only (red dashed; 2017 party-family composition held fixed, actual rates used) and compositional only (blue dotted; 2017 rates held fixed, actual composition used)---with the 2017 baseline shown in gray. \textit{Right}: Annual decomposition of change from the 2017 baseline into behavioral (red), compositional (blue), and interaction (gray) components. Behavioral change accounts for 81--94\% of the total rise, averaging approximately 87\% across 2018--2022. Numbers inside bars show the percentage of total change attributable to behavioral change.}
\label{fig:decomp}
\end{figure}

Behavioral change dominates throughout the period (Figure~\ref{fig:decomp}). From 2018 to 2022, it accounts for 81--94\% of the total increase, averaging approximately 87\%; compositional change accounts for approximately 13\%, with the interaction contributing only a small residual. The rise in incivility is therefore not primarily a consequence of already-uncivil parties gaining prominence, but reflects parties themselves becoming more uncivil.

This result provides the first indication of a broader \textit{mainstreaming of incivility}: the aggregate increase is driven mainly by changing behavior within the party system rather than by the growing weight of actors already located at its more uncivil margins. A party-level decomposition, which rules out compositional changes within party families, yields the same conclusion (Supplementary Information Section~3).

\subsection*{The mainstream becomes more uncivil}

\begin{figure}[htbp]
\centering
\includegraphics[width=0.80\textwidth]{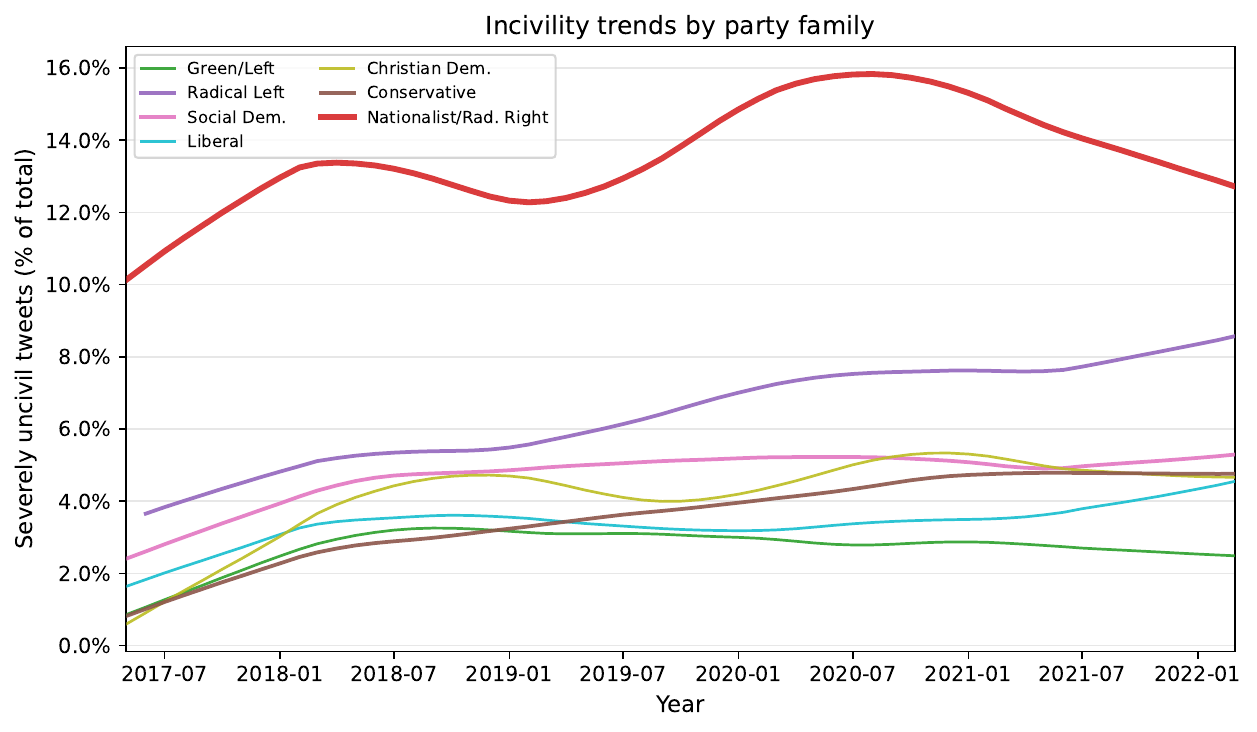}
\caption{\textbf{The mainstream becomes more uncivil.} LOWESS-smoothed severe-incivility rates by party family, 2017--2022. Nationalist and radical-right parties begin at approximately 10.1\%, rise to a peak of $\approx$15.8\% in mid-2020, and partially recede to $\approx$12.7\% by 2021--2022---still above their 2017 level. Other families rise more steadily: radical-left parties from 3.4\% to 8.8\%, conservatives from 0.9\% to 4.4\%, and social democrats from 2.6\% to 5.6\%. Nationalist and radical-right parties remain the most uncivil throughout, but the gap narrows as other party families become steadily more uncivil.}
\label{fig:familytrends}
\end{figure}

Party-family trends show where this behavioral change occurs (Figure~\ref{fig:familytrends}). Nationalist and radical-right parties begin the period with by far the highest rates of severe incivility, at approximately 10.1\% in mid-2017---several times the aggregate average. Their incivility rises sharply to a peak of approximately 15.8\% in mid-2020 before partially receding to 12.7\% by 2021--2022. They nevertheless end the period approximately 2.6 percentage points above where they began.

Other party families follow a different trajectory. Their rise is steadier and largely monotonic: radical-left parties increase from 3.4\% to 8.8\% ($+$5.4~pp), conservative parties from 0.9\% to 4.4\% ($+$3.5~pp), social-democratic parties from 2.6\% to 5.6\% ($+$3.0~pp), and green/left parties from 0.9\% to 3.4\% ($+$2.5~pp). While nationalist and radical-right parties thus remain the most uncivil throughout, the gap narrows as their 2020 peak partially recedes while other party families become steadily more uncivil.

The pattern is consistent with a mainstreaming of incivility, but does not represent a convergence to parity. Aggressive rhetoric is becoming more common across the party system, but nationalist and radical-right parties remain substantially more uncivil than the mainstream at every point in the series.

\subsection*{Populism is the strongest party-level predictor of incivility}

\begin{figure}[htbp]
\centering
\includegraphics[width=0.75\textwidth]{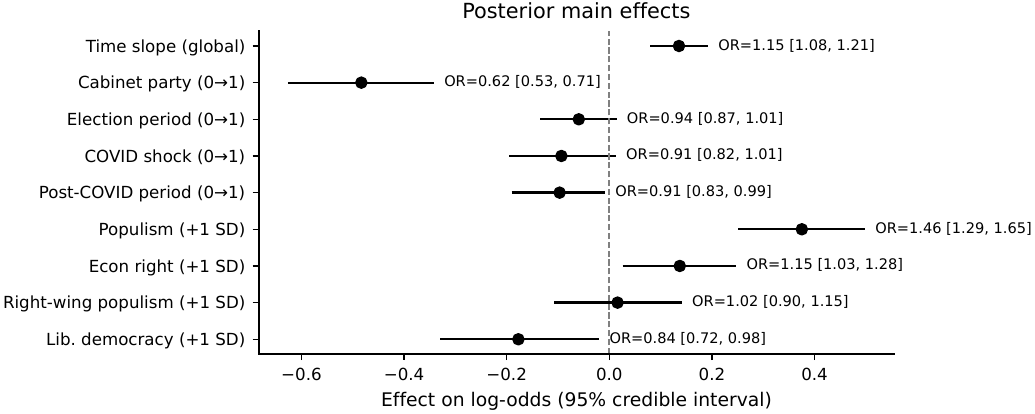}
\caption{\textbf{Main model effect estimates.} Posterior medians and 95\% credible intervals for all model predictors, shown on the log-odds scale, with odds ratios and credible intervals annotated at right. Populism is the strongest party-level predictor (OR~=~1.46), whereas cabinet membership is associated with substantially lower incivility (OR~=~0.62). The time slope confirms a robust upward trend (OR~=~1.15 per SD). The COVID-19 shock estimate is below one (OR~=~0.91), although its credible interval includes 1 for severe incivility. Liberal democracy is negatively associated with severe incivility (OR~=~0.84). $N = 10{,}445$ party--country--month observations, 243 parties, 26 countries.}
\label{fig:forest}
\end{figure}

Populism is the strongest party-level predictor in the main model (Figure~\ref{fig:forest}; full estimates in Extended Data Table~1). A one-standard-deviation increase in a party's V-Party populism score is associated with 46\% higher odds of severe incivility (OR~=~1.46, 95\%~CI: 1.29--1.65). Economic right-wing position is also positively associated with incivility (OR~=~1.15, 95\%~CI: 1.03--1.28), but the interaction between populism and economic ideology is null (OR~=~1.02, 95\%~CI: 0.90--1.15). The association between populism and incivility is therefore not confined to the economic right.

\begin{figure}[htbp]
\centering
\includegraphics[width=0.65\textwidth]{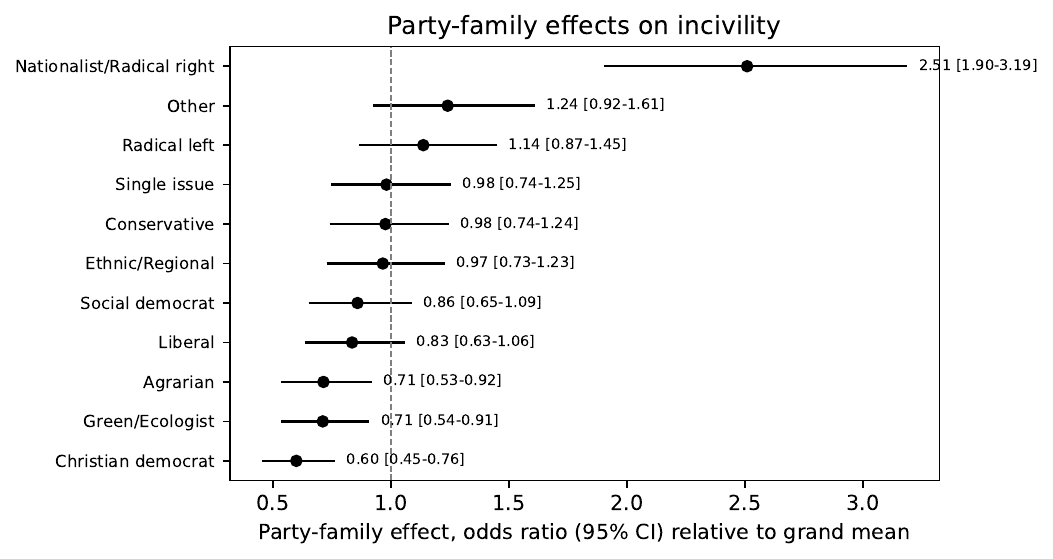}
\caption{\textbf{Party-family effects on incivility.} Posterior medians and 95\% credible intervals of partially pooled party-family effects, expressed as odds ratios relative to the grand mean, from a model replacing the continuous populism predictor with party-family intercepts and controlling for country and time. Nationalist and radical-right parties are a clear outlier (OR~=~2.51, 95\%~CI: 1.90--3.19), with an effect approximately twice that of the next-highest family, the radical left (OR~=~1.14). Christian democratic parties are the least uncivil (OR~=~0.60).}
\label{fig:partyfam}
\end{figure}

This does not imply that all populist party families are equally uncivil in absolute terms: a complementary model replacing the continuous populism predictor with partially pooled party-family effects identifies nationalist and radical-right parties as a clear outlier (Figure~\ref{fig:partyfam}). Their estimated odds of severe incivility are approximately two and a half times the grand mean (OR~=~2.51, 95\%~CI: 1.90--3.19), compared with OR~=~1.14 (95\%~CI: 0.87--1.45) for the next-highest family, the radical left. The two results are compatible: populism is associated with greater incivility across ideological positions, while nationalist and radical-right parties remain distinctive in their absolute levels.

Cabinet membership is associated with substantially lower incivility: governing parties produce content with 38\% lower odds of severe incivility than opposition parties (OR~=~0.62, 95\%~CI: 0.53--0.71), consistent with a strategic account in which opposition parties face stronger incentives for norm-breaking rhetoric---though this interpretation remains correlational. 

A populism~$\times$~opposition interaction model (Extended Data Figure~1) shows these associations reinforce one another: the interaction is OR~=~1.61 (95\%~CI: 1.26--2.06, $\Delta p = +2.3$~pp), so the populism--incivility gap is substantially wider when populist parties are in opposition. A null populism~$\times$~time interaction (OR~=~0.990, 95\%~CI: 0.979--1.001; Extended Data Figure~4) indicates that the temporal rise does not reflect a strengthening of the populism--incivility link: party families across the spectrum are becoming more uncivil, not populist ones alone.

\subsection*{Issue content shapes incivility, but the populism effect is not topic-specific}

One possible explanation for the party-level differences is issue selection. Populist parties may appear more uncivil because they concentrate on migration, nationalism, crime, or identity---topics that themselves generate more conflict. If so, the association between populism and incivility should be concentrated in these domains.

\begin{figure}[htbp]
\centering
\includegraphics[width=0.65\textwidth]{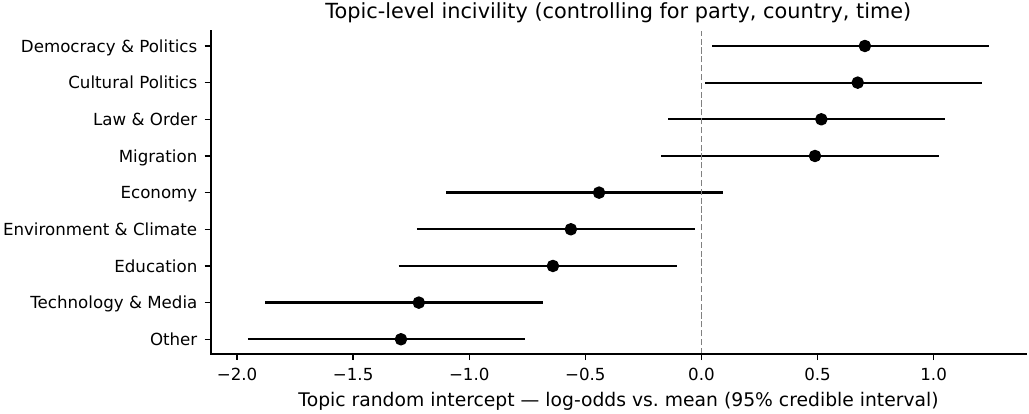}
\caption{\textbf{Issue-area differences in incivility.} Posterior medians and 95\% credible intervals of topic random intercepts from the issue-area model, expressed as odds ratios relative to the grand mean and sorted by posterior median. Democracy \& Politics (OR~=~2.03) and Cultural Politics (OR~=~1.97) are the most uncivil topics; Technology \& Media (OR~=~0.30) and Education (OR~=~0.52) are the least uncivil substantive topics. The residual ``Other'' category is marginally lower (OR~=~0.28) but does not represent a coherent issue area. Estimates account for party characteristics, country, and time. $N = 82{,}015$ observations across 9 topic categories.}
\label{fig:topics}
\end{figure}

Issue content strongly predicts the baseline level of incivility (Figure~\ref{fig:topics}). In the model estimated on 82,015 party--country--month--topic observations across nine categories, Democracy \& Politics (OR~=~2.03), Cultural Politics (OR~=~1.97), Law \& Order (OR~=~1.68), and Migration (OR~=~1.62) are markedly more uncivil than average. Technology \& Media (OR~=~0.30), Education (OR~=~0.52), Environment \& Climate (OR~=~0.56), and Economy (OR~=~0.64) are substantially less so. Incivility is therefore strongly structured by issue content, with disputes over political legitimacy and identity producing far higher rates than more technical policy domains.

The populism association, however, is not concentrated in any particular topic. All nine topic-specific populism slopes are positive and broadly similar in magnitude, although their individual credible intervals include zero (Extended Data Figure~3). The pooled slope is also imprecisely estimated in this more finely stratified model (OR~=~1.21, 95\%~CI: 0.85--1.72).

The relevant comparison is between topics: neither Migration nor Cultural Politics---nor any other category---shows a distinctly stronger populism slope than the rest. This is evidence of no detectable concentration, not proof that the association is identical across topics; smaller differences cannot be excluded. But the results provide no indication that the precisely estimated main-model association between populism and incivility (OR~=~1.46) is driven by a narrow set of high-conflict issues.

\subsection*{Stronger liberal democracy is associated with lower severe incivility}



Liberal democracy is negatively associated with severe incivility: a one-standard-deviation increase in V-Dem's liberal democracy index corresponds to 16\% lower odds of severe incivility (OR~=~0.84, 95\%~CI: 0.72--0.98), based on cross-national variation after accounting for party-level characteristics.

Notably, this association is concentrated at the severe end of the distribution. When the outcome is expanded to include any incivility ($\geq 1$), the coefficient is null (OR~=~1.01, 95\%~CI: 0.90--1.14). This threshold difference is consistent with stronger democratic institutions, parliamentary norms, and social sanctions constraining the most extreme rhetorical violations more than ordinary political sharpness, although the observational design does not establish the mechanism.

A populism~$\times$~liberal-democracy model (Extended Data Figure~2) yields a positive interaction (OR~=~1.17, 95\%~CI: 1.01--1.34): the incivility gap between more and less populist parties is somewhat wider in stronger democracies. One possible interpretation is that populist parties stand out more sharply where the surrounding party system is less uncivil.

A complementary within-country analysis points in the same direction. Country-years in which liberal democracy is weaker than that country's own average are associated with higher incivility (OR~=~0.969 per SD increase in liberal democracy, 95\%~CI: 0.954--0.985; Supplementary Information Section~1). The cross-national association is therefore accompanied by a similar, though modest, within-country pattern.

\subsection*{Robustness}

The core findings hold across alternative operationalizations and measurement instruments. First, redefining the outcome as any incivility ($\geq 1$) rather than severe incivility ($\geq 2$) reproduces the main party-level structure (Supplementary Information Section~3; Table~\ref{tab:si_any}). Populism remains positively associated with incivility (OR~=~1.25, 95\%~CI: 1.12--1.39), and governing parties remain substantially less uncivil than opposition parties (OR~=~0.49, 95\%~CI: 0.43--0.55). The time slope also remains positive (OR~=~1.06, 95\%~CI: 1.01--1.10), although it is attenuated relative to the severe-incivility estimate, consistent with the rise being strongest for the most hostile content. The COVID-19 decline is likewise credible at this lower threshold (OR~=~0.87, 95\%~CI: 0.81--0.93).

Second, a cross-instrument analysis using Perspective API toxicity scores as the dependent variable provides convergent evidence for the party-level findings (Supplementary Information Section~2). The associations with populism and cabinet status replicate directionally, with credible intervals excluding the null, on this independent instrument. We do not use Perspective to assess the longitudinal or cross-national results because its scores are temporally non-stationary owing to undisclosed model retrainings~\citep{hartmann2026bye} and because it supports only a subset of the languages in our sample. Countries whose politicians tweet primarily in unsupported languages are therefore represented only by their supported-language content, which may not be representative. We consequently treat this analysis as a within-language, cross-sectional validation of the party-level structure.

Third, we address concerns about the replicability of measurements produced by proprietary large language models~\citep{barrie2026replication}. Although the exact GPT-4.1 model may not remain available indefinitely, our measurement procedure is not tied to that model: we share the human-coded validation sample and scoring rubric, so the classification can in principle be reproduced with any sufficiently capable model that achieves comparable performance against the same human benchmark. To make the measurement step fully rerunnable with an openly available model, we additionally train and publicly release a downloadable XLM-RoBERTa-large classifier and rebuild the full panel using its predictions (Supplementary Information Section~6). The open classifier achieves strong performance against the human-coded benchmark, although somewhat below GPT-4.1, and we therefore retain GPT-4.1 as the primary measure. Importantly, the open classifier reproduces the central findings: the time trend (OR~=~1.09, 95\%~CI: 1.04--1.15), cabinet association (OR~=~0.61, 95\%~CI: 0.54--0.70), and populism association (OR~=~1.31, 95\%~CI: 1.18--1.45) all remain credible. The liberal democracy coefficient remains negative (OR~=~0.89), although its interval includes 1 (95\%~CI: 0.75--1.07), consistent with the lower measurement precision of the open classifier attenuating this subtler cross-national association.

\section{Discussion}\label{sec:discussion}
Leveraging the largest cross-national longitudinal dataset on elite political incivility assembled to date, this paper makes three advances. We document a robust rise in severe incivility across 26 democracies between 2017 and 2022, show that this rise is overwhelmingly behavioral rather than compositional, and identify populism, opposition status, and democratic quality as its strongest correlates. Together, these findings describe a mainstreaming of incivility: nationalist and radical-right parties remain the most uncivil party family throughout the period, but the gap between them and the rest of the political spectrum is narrowing---not because the radical right is moderating, but because everyone else is catching up.

The central question this raises is what drives the rise across party families. Opposition status, populism, and democratic quality explain \textit{who} is relatively more uncivil at a given moment, but none of them explains why radical-left, conservative, and social-democratic parties would all become steadily more aggressive over the same five years. Two non-mutually exclusive mechanisms plausibly combine to produce this pattern. The first is competitive rhetorical adaptation: when a radical challenger gains visibility and votes by breaking discursive norms, mainstream competitors face pressure to match its register or lose ground in framing and attention. The timing in our data is consistent with this reading---the radical right's incivility is already elevated at the start of the series and rises further before the mainstream's own increase gathers pace---though our design cannot establish that one caused the other. The second is a structural pressure operating on all parties regardless of ideology: platforms reward engagement, and moral-emotional, outrage-laden language reliably generates it~\citep{brady2017emotion,tornberg2022digital}. Under this account, mainstream parties are not converging on the radical right specifically so much as every party is being pulled toward whatever rhetorical register the attention economy subsidizes, with the radical right simply arriving there first because confrontation was already central to its appeal. Distinguishing imitation from shared incentive is beyond what our data can do, but the two are not competing explanations: social media may define the selection environment while the radical right's example creates competitive pressure that pushes other parties to adapt.

The government--opposition results point to a second, complementary mechanism operating at the level of individual parties. Cabinet parties are substantially less uncivil than opposition parties (OR~=~0.62), and this gap is considerably wider for populist parties specifically (populism~$\times$~opposition interaction, OR~=~1.61, 95\%~CI: 1.26--2.06). This is consistent with the inclusion-moderation thesis, under which the responsibilities of governing---coalition management, institutional accountability, the reputational cost of destabilizing rhetoric while in office---constrain a confrontational style that carries little such cost in opposition~\citep{akkerman2016radical}. It also offers a more specific reading of the radical right's partial post-2020 retreat: rather than a general softening of tone, some of that decline may track growing government participation among radical-right parties in our sample.

Populism's association with incivility is not concentrated in migration, cultural politics, or any other single issue area; all nine topic-specific slopes point in the same direction and are of broadly similar magnitude. This weighs against a simple issue-selection account in which populist parties merely appear more uncivil because they talk more about inherently conflictual topics. It is more consistent with populism operating as a thin-centred ideology~\citep{mudde2004populist}---a Manichean frame of a virtuous people against a corrupt elite that can attach to essentially any policy domain. On this reading, incivility is not a rhetorical accessory that populist parties happen to deploy more often, but close to intrinsic to how the antagonism at the core of populist communication gets expressed, regardless of subject matter---though our topic-level estimates are imprecise enough that smaller differences across issues cannot be ruled out.

Stronger liberal democracy is associated with lower severe incivility, both cross-nationally and, more modestly, within countries over time, and this association is concentrated specifically at the severe end of the distribution rather than at the threshold of ordinary political sharpness. This pattern is consistent with formal parliamentary rules, professional norms, and social sanction constraining the most egregious violations while leaving everyday partisan combat largely untouched, though our data cannot identify the mechanism directly. Nor can they establish the direction of causality. Democratic erosion may loosen constraints on aggressive elite rhetoric, while elite incivility may itself contribute to erosion by intensifying polarization and increasing tolerance for anti-democratic conduct~\citep{gervais2019rousing,gessler2025newregime,orhan2022relationship}. Both the cross-national and within-country results are consistent with a mutually reinforcing relationship between the two, without allowing us to adjudicate which, if either, comes first.

Several limitations bear noting. Our measure captures one dimension of communicative norm violation---hostility, disrespect, personal attack---and should not be equated with democratic degradation more broadly. Twitter/X is not a neutral mirror of political discourse: it privileges particular politicians, audiences, and communicative styles, and tweeting parliamentarians may not represent elite political communication as a whole. Cross-linguistic comparability remains an inherent challenge even with extensive human validation. And our design, however extensively triangulated, is observational: it identifies robust associations and a clear temporal pattern, not causal mechanisms.

These caveats do not soften the central finding: elite political incivility nearly doubled across 26 democracies in five years, and the increase is driven mainly by parties across the spectrum becoming more aggressive, not by already-uncivil actors gaining prominence. Nationalist and radical-right parties remain the most uncivil political family throughout, but aggressive and demeaning rhetoric is becoming markedly more common well beyond that margin. Exposure to uncivil elite communication lowers institutional trust~\citep{mutz2005new}, increases citizens' own aggressive political intentions~\citep{gervais2017more}, and raises tolerance for anti-democratic behavior~\citep{gervais2019rousing}. A near-doubling of elite incivility across 26 democracies, driven by change throughout the party system rather than at its margins, represents a substantial and broadly shared shift in the rhetorical environment democratic citizens now encounter.

\section{Methods}\label{sec:methods}

\subsection*{Data}

\paragraph{Tweets.}
We assembled a corpus of tweets sent by national parliamentarians in 26 democracies between May 2017 and March 2022: Australia, Austria, Belgium, Canada, Denmark, Finland, France, Germany, Greece, Iceland, Ireland, Italy, Latvia, Luxembourg, Malta, the Netherlands, New Zealand, Norway, Poland, Slovenia, Spain, Sweden, Switzerland, Turkey, the United Kingdom, and the United States~\citep{van2020twitter}. The corpus comprises 26,683,450 tweets, collected under Twitter's Academic Research access tier prior to the platform's data-policy changes.

\paragraph{Incivility labelling.}
We define \textit{political incivility} as hostile, disrespectful, or contemptuous political language, following established usage~\citep{mutz2005new,theocharis2020dynamics}; the concept is deliberately narrower than the full set of democratic norm violations, and does not encompass misinformation, affective polarization, or hate speech as such. We reserve the term ``toxicity'' for the Perspective API instrument (Supplementary Information Section~2), which operationalizes a broader, partly overlapping construct.

Each tweet was scored on a four-point ordinal scale (0 = civil to 3 = extremely uncivil) using \texttt{gpt-4.1} (snapshot \texttt{gpt-4.1-2025-04-14}) via the OpenAI Batch API at temperature 0, an approach validated for political and social-scientific text classification in prior work~\citep{gilardi2023chatgpt,tornberg2023chatgpt}. Tweets were scored in their original language without translation; URLs, @-mentions, and hashtags were retained as they appeared; retweets were excluded. The model received only tweet text, with no party, politician, or country metadata, precluding any influence of prior expectations about a party's reputation on its scores. The full prompt and rubric appear in Supplementary Information Section~4. The primary dependent variable is the per-cell count of tweets scoring $\geq 2$ (``severe incivility,'' i.e.\ very or extremely uncivil).

\paragraph{Validation.}
Four trained coders collectively annotated a stratified random sample of 1,200 tweets (1,195 successfully coded), working blind to the model's scores. To test the classifier at its most consequential decision point, the sample was boundary-weighted around the 1/2 threshold used in the main analysis (450 tweets at score 1, 450 at score 2, 150 each at scores 0 and 3) and stratified by language family. The team was assembled for native-language coverage: 75\% of items were coded by a coder working directly in the tweet's original language, including all Turkish and Greek items by native speakers; the remaining 25\%---chiefly Polish, Finnish, and Latvian, for which no native coder was available---used DeepL assistance.

Because the boundary-weighted design deliberately oversamples ambiguous cases, it understates performance on the unambiguous majority; we therefore report both raw agreement and agreement reweighted to the true corpus distribution (83\% score-0, 12\% score-1, 4.5\% score-2, 0.5\% score-3). Raw quadratic-weighted agreement is $\kappa = 0.74$, rising to $\kappa = 0.80$ under reweighting---at or above the inter-coder agreement of the two primary human coders ($\kappa = 0.77$, $n = 197$), indicating that the classifier achieves agreement comparable to the inter-coder benchmark for this inherently subjective construct. Agreement is comparable across language families (Romance 0.84, Germanic 0.83, Turkish 0.80, English 0.76), with the Greek specialist sample the one exception ($\kappa = 0.63$, $n = 46$; an acknowledged limitation). A total of 97\% of classifications fall within one ordinal category of the human label.

At the binary $\geq 2$ threshold, the classifier is conservative rather than over-inclusive: precision is 0.84 and recall 0.65, so it misses some severe content rather than over-flagging civil content. This matters for interpretation. Provided the resulting misclassification is approximately non-differential with respect to party, country, and time, it attenuates rather than inflates estimated effects, making our estimates conservative. We cannot prove non-differentiality, but we test its most plausible failure modes directly and find no evidence of bias in the direction that would concern us: DeepL-assisted and natively-coded items show near-identical agreement (Supplementary Information Section~4); country-level classifier agreement is not credibly associated with estimated incivility; and disagreement for radical-right and high-populism tweets is driven by \textit{under}-classification, not over-classification, so any residual bias understates rather than manufactures the party-level effects we report (Supplementary Information Section~4). The structural findings additionally replicate on the Perspective API, an entirely independent instrument (Supplementary Information Section~2).

\paragraph{Political party data.}
Party-level covariates were drawn from V-Party (V-Dem v2)~\citep{lindberg2014v} and ParlGov 2024~\citep{doring2024parlgov}, linked via the Party Facts crosswalk~\citep{bederke2022partyfacts}. The populism index ($v2xpa\_popul$, range 0--1) captures the degree to which a party combines anti-establishment rhetoric with appeals to popular sovereignty. Economic left--right position ($v2pariglef$) was taken from V-Party and, for parties it does not cover, supplemented with ParlGov scores rescaled to the same metric (8.9\% missing after filling). Cabinet membership was derived from ParlGov records as each party's monthly fraction of time in government. Country-level liberal democracy was measured by V-Dem's liberal democracy index ($v2x\_libdem$), entered as a country-level mean (averaged across the study period) and z-scored as a cross-sectional predictor in the main model.

\paragraph{Topic classification.}
Each tweet was assigned to one of nine topic categories using zero-shot classification with \texttt{gpt-4.1-mini} (snapshot \texttt{gpt-4.1-mini-2025-04-14}). The nine categories collapse an initial twelve-category scheme along three conceptually overlapping boundaries---Economy (Economic + Welfare), Cultural Politics (Identity + Nationalism), and Law \& Order (Crime + Security)---with the other six unchanged. Validation used a 600-tweet sample balanced across categories and language families; two coders annotated overlapping subsets so every item received at least one label, with 204 items double-coded. Inter-coder agreement is $\kappa = 0.85$. Against the gold-standard subset on which both coders agreed ($n = 173$), classifier agreement is $\kappa = 0.80$ (82.7\% exact); benchmarked against the full double-coded set including the 31 disagreement cases, it is $\kappa = 0.78$ and $0.71$ against the two coders individually---modestly lower, as expected when ambiguous items are retained (Supplementary Information Section~4).

\subsection*{Panel construction}
We aggregated tweets to a \textit{party--country--month} panel, retaining cells with at least 10 tweets for stable rate estimates; results are robust to thresholds of 10, 30, and 50 tweets per cell (Supplementary Information Section~3). The main panel comprises 10,445 observations across 243 parties, 26 countries, and 59 months (13,790,734 tweets). Stratifying additionally by topic yields the 82,015-observation issue-area panel. All time-varying predictors (cabinet fraction, election-period fraction, COVID-19 shock, post-COVID period) were constructed as monthly values at the panel's resolution. 

\subsection*{Hierarchical Bayesian beta-binomial model}
Let $y_{ict}$ be the count of severe-incivility tweets and $n_{ict}$ the total tweet count for party $i$ in country $c$ at month $t$. We model
\begin{align}
y_{ict} &\sim \mathrm{BetaBinomial}\!\left(n_{ict},\; \mu_{ict}\kappa,\; (1-\mu_{ict})\kappa\right) \\
\mathrm{logit}(\mu_{ict}) &= \alpha + \boldsymbol{x}_{ict}^\top\boldsymbol{\beta}_w + \gamma_t + u_c + v_i + (\beta_t + b_c)\,s_t,
\end{align}
where $\boldsymbol{x}_{ict}$ are within-observation predictors (cabinet fraction, election fraction, COVID shock, post-COVID period), $s_t$ is standardized time, $\gamma_t$ are i.i.d.\ month random effects, and $\kappa$ is a beta-binomial overdispersion parameter (prior Exponential(1/50)) that lets the model accommodate variance beyond the binomial and so guards against overconfident intervals. Rather than treating the country intercept $u_c$, country time-slope deviation $b_c$, and party intercept $v_i$ as free random effects, we model them as functions of the country- and party-level predictors, with a residual term absorbing remaining heterogeneity:
\begin{align}
u_c &= \gamma^{u}_{\text{libdem}}\,\text{libdem}_c + \tilde{u}_c \\
b_c &= \gamma^{b}_{\text{libdem}}\,\text{libdem}_c + \tilde{b}_c, \qquad
(\tilde{u}_c, \tilde{b}_c) \sim \mathrm{Normal}\ \text{(LKJ}(\eta{=}2)\text{ correlated)} \\
v_i &= \gamma_{\text{pop}}\,\text{populism}_i + \gamma_{\text{econ}}\,\text{econ}_i + \gamma_{\text{pop}\times\text{econ}}(\text{populism}_i \times \text{econ}_i) + \tilde{v}_i, \quad \tilde{v}_i \sim \mathrm{Normal}(0,\sigma_v).
\end{align}
Here $\text{libdem}_c$, $\text{populism}_i$, and $\text{econ}_i$ are z-scored. This structure is deliberate: modelling the group-level intercepts as regressions on covariates, with partial-pooling residuals, lets each country- and party-level effect ($\gamma^{u}_{\text{libdem}}$, $\gamma_{\text{pop}}$, $\gamma_{\text{econ}}$, $\gamma_{\text{pop}\times\text{econ}}$) be estimated while the residuals $\tilde{u}_c,\tilde{b}_c,\tilde{v}_i$ absorb idiosyncratic between-group variation, accounting for residual country- and party-level heterogeneity. These coefficients are exactly the ``Liberal democracy,'' ``Populism index,'' ``Econ left--right,'' and ``Right-wing populism interaction'' effects in Figure~\ref{fig:forest} and Extended Data Table~1. The model additionally estimates $\gamma^{b}_{\text{libdem}}$ in the country time-slope equation, which allows the rate of change to vary with liberal democracy; this parameter serves as a structural control and is not separately reported. Priors are weakly informative throughout (coefficients $\mathrm{Normal}(0,0.25)$; scale parameters Exponential(2)); $(\tilde{u}_c,\tilde{b}_c)$ use an LKJ-Cholesky covariance and $\tilde{v}_i$ a non-centred parameterization for sampling efficiency.

We fit the model with the NumPyro NUTS sampler via PyMC (4 chains $\times$ 2,000 draws after 2,000 tuning steps; target acceptance 0.99). All reported models converged with zero divergences and $\hat{R} \leq 1.01$ on all parameters. Posterior summaries are medians with 95\% equal-tailed credible intervals.

\subsection*{Shift-share decomposition}
To separate behavioral from compositional contributions to the trend, we apply a Blinder--Oaxaca-style shift-share decomposition~\citep{oaxaca1973male}. With $w_{ft}$ the tweet-volume share of family $f$ at month $t$ and $r_{ft}$ its incivility rate, the observed aggregate is $R_t = \sum_f w_{ft} r_{ft}$. Holding composition at 2017 levels isolates behavioral change, $R^A_t = \sum_f w_{f,2017}\, r_{ft}$; holding rates at 2017 levels isolates compositional change, $R^B_t = \sum_f w_{ft}\, r_{f,2017}$. The behavioral and compositional components are $R^A_t - R_{2017}$ and $R^B_t - R_{2017}$, with the interaction as residual. A party-level version (Supplementary Information Section~3) confirms the result is not an artifact of within-family composition.

\subsection*{Issue-area model}
The issue-area model extends the main model with topic random intercepts $a_{t'}$ and partially pooled topic-specific populism slopes $b^{pop}_{t'}$:
\begin{align}
\mathrm{logit}(\mu_{ictt'}) &= \cdots + a_{t'} + b^{pop}_{t'}\, z^{pop}_i, \\
a_{t'} \sim \mathrm{Normal}(0,\sigma_{topic}), \qquad
b^{pop}_{t'} &\sim \mathrm{Normal}(\mu^{pop},\sigma^{pop}),
\end{align}
where $t'$ indexes the nine topics and $z^{pop}_i$ is party populism (z-scored). Partial pooling of the slopes is the appropriate test for topic-specific instrumentalization: it shrinks imprecise per-topic estimates toward the common mean, so an outlying topic must be supported by the data to appear as one. Country effects use a non-LKJ structure for tractability at 82,015 observations.

\subsection*{Robustness specifications}
Three interaction variants of the main model are reported in Extended Data, each adding a single cross-level interaction term: populism $\times$ opposition (Figure~1), populism $\times$ liberal democracy (Figure~2), and populism $\times$ time (Figure~4). Further checks appear in Supplementary Information: a within-country democratic-backsliding model (Section~1); a Perspective API beta-regression providing cross-instrument validation (Section~2); threshold sensitivity at the any-incivility ($\geq 1$) level (Section~3); full classifier validation with language-specific statistics (Section~4); leave-one-country-out refits for the four countries with the most leverage on the liberal democracy coefficient (Section~5); and a full replication with an open-source XLM-RoBERTa-large classifier trained on the GPT-4.1 labels (Section~6).

\subsection*{Code and data availability}
All analysis code, aggregated data, and the human-coded validation sample are available at \url{https://osf.io/fg46r/}. The repository includes the aggregated party--country--month panel with per-score incivility counts ($n_{\text{score0}}$--$n_{\text{score3}}$) alongside the thresholded indicators used in the analysis, allowing the dependent variable to be re-thresholded without the tweet-level scores; tweet IDs with their GPT-4.1 incivility labels for all 13.8~million classified tweets, so that the full aggregation and every reported model can be reproduced without re-running the classifier; and the human-coded validation sample (1,195 tweets, tweet IDs and per-coder scores). Consistent with Twitter/X terms of service, raw tweet text is not redistributed but can be re-hydrated from the IDs with API access. The classification prompt, scoring rubric, and model identifier are documented in Supplementary Information Section~4.

\clearpage
\section*{Extended Data}

\begin{table}[ht]
\centering
\caption{\textbf{Extended Data Table~1: Main model effects on political incivility.} Hierarchical Bayesian beta-binomial model. Dependent variable: count of severe-incivility tweets (GPT-4.1 score $\geq 2$) per party--country--month cell. Odds ratios (OR) and 95\% credible intervals (CI) from the posterior ($n = 10{,}445$ observations, 243 parties, 26 countries). Intervals excluding 1.0 are reported in the text as credible associations. Zero divergences in all four chains; values match Figure~\ref{fig:forest}.}
\label{tab:effects}
\begin{tabular}{lrrr}
\toprule
\textbf{Predictor} & \textbf{OR} & \textbf{95\% CI} & \textbf{$\Delta p$ (pp)} \\
\midrule
\multicolumn{4}{l}{\textit{Time and event effects}} \\
\quad Time slope (per SD) & 1.15 & [1.08,\ 1.21] & $+$0.58 \\
\quad COVID-19 shock & 0.91 & [0.82,\ 1.01] & $-$0.36 \\
\quad Post-COVID period & 0.91 & [0.83,\ 0.99] & $-$0.37 \\
\quad Election period & 0.94 & [0.87,\ 1.01] & $-$0.23 \\
\addlinespace
\multicolumn{4}{l}{\textit{Party-level predictors}} \\
\quad Cabinet party (vs.\ opposition) & 0.62 & [0.53,\ 0.71] & $-$1.56 \\
\quad Populism index (+1 SD) & 1.46 & [1.29,\ 1.65] & $+$1.79 \\
\quad Economic right (+1 SD) & 1.15 & [1.03,\ 1.28] & $+$0.59 \\
\quad Right-wing populism interaction & 1.02 & [0.90,\ 1.15] & $+$0.07 \\
\addlinespace
\multicolumn{4}{l}{\textit{Country-level predictor}} \\
\quad Liberal democracy (+1 SD) & 0.84 & [0.72,\ 0.98] & $-$0.65 \\
\bottomrule
\end{tabular}
\smallskip\\
\footnotesize{$\Delta p$ is the median posterior change in predicted incivility rate, in percentage points, evaluated at the grand mean of all other predictors.}
\end{table}

\begin{figure}[ht]
\centering
\includegraphics[width=0.85\textwidth]{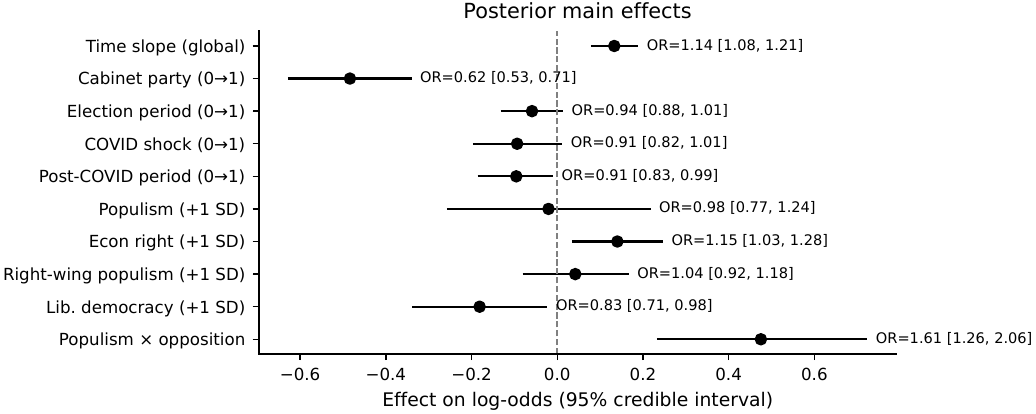}
\caption{\textbf{Extended Data Figure~1: Populism $\times$ opposition interaction model.} Forest plot from a variant of the main model adding a cross-level interaction between party populism (z-scored) and monthly opposition fraction ($1 -$ cabinet fraction). The positive and credible interaction term (OR~=~1.61, 95\%~CI: 1.26--2.06, $\Delta p = +2.3$~pp) indicates that the populism--incivility association is amplified when populist parties are in opposition. All other effects remain consistent with the main model.}
\label{fig:extdata_opp}
\end{figure}

\begin{figure}[ht]
\centering
\includegraphics[width=0.85\textwidth]{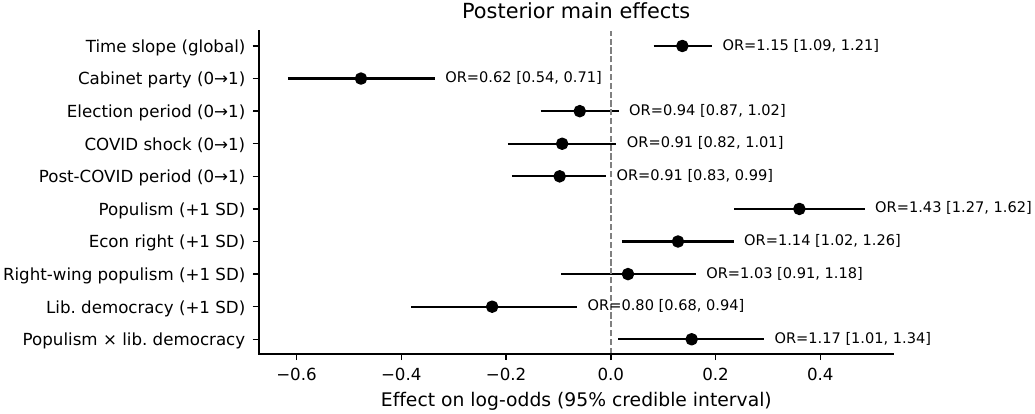}
\caption{\textbf{Extended Data Figure~2: Populism $\times$ liberal democracy interaction model.} Forest plot from a variant of the main model adding a cross-level interaction between party populism (z-scored) and country-level V-Dem liberal democracy (z-scored). The positive and credible interaction term (OR~=~1.17, 95\%~CI: 1.01--1.34) indicates that the populism--incivility gap is modestly wider in stronger democracies, where populist parties may stand out more sharply against a civil mainstream. All other effects remain consistent with the main model.}
\label{fig:extdata_libdem}
\end{figure}

\begin{figure}[ht]
\centering
\includegraphics[width=0.85\textwidth]{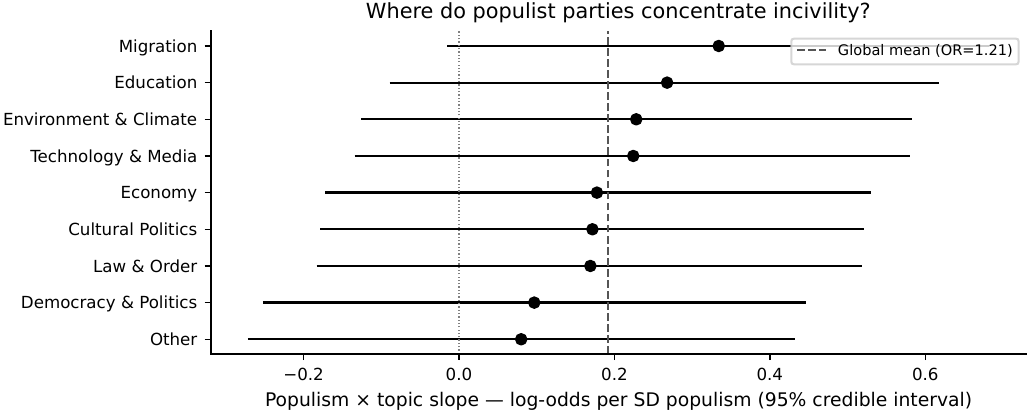}
\caption{\textbf{Extended Data Figure~3: Populism--incivility association by topic.} Posterior medians and 95\% credible intervals of the populism slope for each of the nine topic categories, estimated from the issue-area model. All nine slopes are positive and of broadly similar magnitude; none has a credible interval clearly excluding zero, and none is distinctly larger than the others. The pooled global slope (OR~=~1.21, 95\%~CI: 0.85--1.72) is itself imprecisely estimated, but the absence of any topic standing out---including Migration and Cultural Politics---indicates that the populism effect is not concentrated in culture-war issues specifically.}
\label{fig:extdata_topicpop}
\end{figure}

\begin{figure}[ht]
\centering
\includegraphics[width=0.85\textwidth]{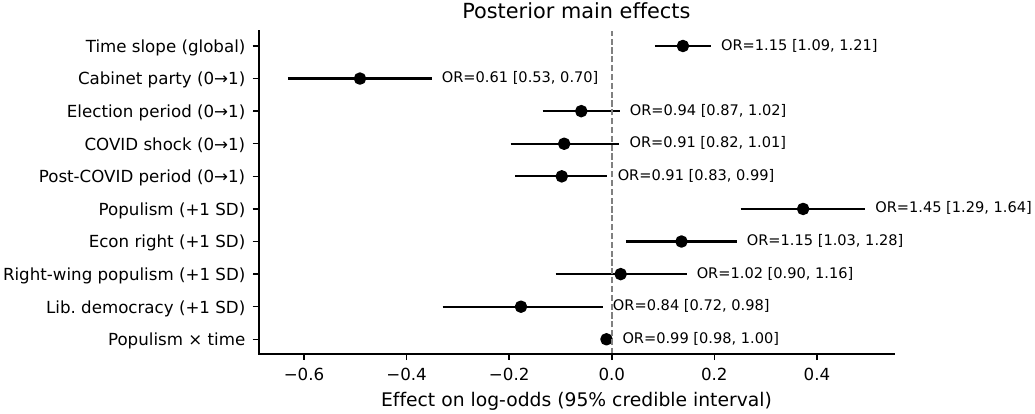}
\caption{\textbf{Extended Data Figure~4: Populism $\times$ time interaction model.} Forest plot from a variant of the main model adding an interaction between party populism (z-scored) and the global standardized time trend, testing whether the populism--incivility association strengthened over the study period. The interaction term is essentially null (OR~=~0.990, 95\%~CI: 0.979--1.001), indicating that the temporal rise does not reflect a widening gap between populist and non-populist parties---consistent with party families across the spectrum becoming more uncivil at broadly comparable rates.}
\label{fig:extdata_poptime}
\end{figure}

\clearpage
\setcounter{table}{0}
\setcounter{figure}{0}
\renewcommand{\thetable}{S\arabic{table}}
\renewcommand{\thefigure}{S\arabic{figure}}
\section*{Supplementary Information}

\subsection*{SI Section 1: Democratic backsliding and within-country variation in liberal democracy}

The main model identifies the liberal democracy effect from cross-national variation: countries with stronger institutions show lower incivility on average, after adjusting for party composition. A complementary question is whether \textit{within-country} changes in democratic quality---year-on-year deterioration relative to a country's own mean---track changes in incivility over time.

We estimated a variant of the main model adding a time-varying predictor for each country-year's deviation from its own mean liberal democracy score ($\Delta\mathrm{libdem}_{ct}$, z-scored within country). This exploits within-country temporal variation in V-Dem scores, which is most pronounced in backsliding countries (e.g.\ Slovenia, range 0.176; Greece, 0.159; Poland, 0.127 over the study period). Mean $\Delta\mathrm{libdem}$ correlates negatively with year ($r = -0.39$), declining from $+0.011$ in 2017 to $-0.012$ in 2022---a modest aggregate democratic decline over the period.

\begin{figure}[ht]
\centering
\includegraphics[width=0.85\textwidth]{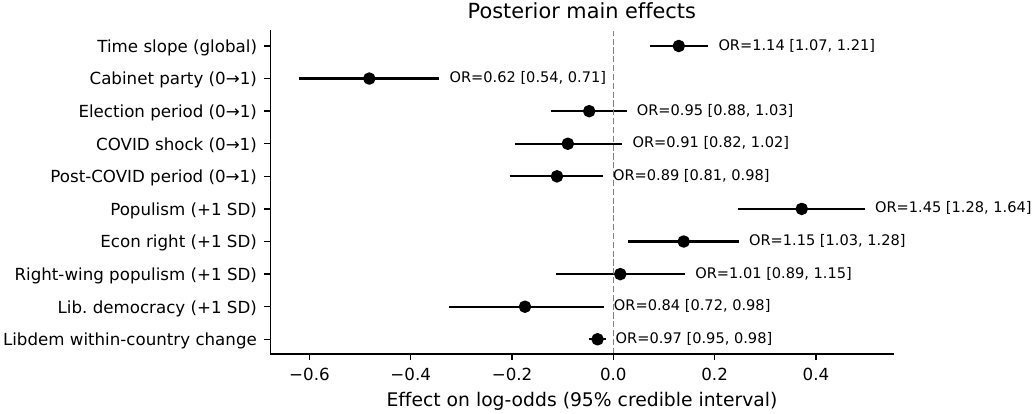}
\caption{\textbf{Democratic backsliding model.} Forest plot of posterior medians and 95\% credible intervals. The bottom row shows the within-country liberal democracy deviation term ($\Delta$libdem); a credible interval below 1.0 indicates that years in which a country's democratic quality falls below its own mean are associated with higher incivility.}
\label{fig:si_libdem_trend}
\end{figure}

The association is credible and in the expected direction (Supplementary Figure~\ref{fig:si_libdem_trend}; Supplementary Table~\ref{tab:si_libdem_trend}): a one-standard-deviation \textit{decline} in a country's liberal democracy relative to its own baseline is associated with approximately 3.2\% higher odds of incivility (OR~=~1.032, 95\%~CI: 1.015--1.048; equivalently, OR~=~0.969 per SD \textit{increase}). The effect is modest in magnitude ($\Delta p \approx -0.12$~pp per SD), consistent with democratic institutions constraining incivility norms while within-country variation over this relatively short window remains limited. All other model effects are stable relative to the main model.

\begin{table}[ht]
\centering
\caption{\textbf{Democratic backsliding model.} Main-model predictors plus within-country liberal democracy deviation ($\Delta$libdem, z-scored). Odds ratios and 95\% credible intervals. $n = 10{,}445$ observations.}
\label{tab:si_libdem_trend}
\begin{tabular}{lrrr}
\toprule
\textbf{Predictor} & \textbf{OR} & \textbf{95\% CI} & \textbf{$\Delta p$ (pp)} \\
\midrule
\multicolumn{4}{l}{\textit{Time and event effects}} \\
\quad Time slope (per SD)         & 1.14 & [1.07,\ 1.21] & $+$0.55 \\
\quad COVID-19 shock               & 0.91 & [0.82,\ 1.02] & $-$0.35 \\
\quad Post-COVID period            & 0.89 & [0.81,\ 0.98] & $-$0.42 \\
\quad Election period              & 0.95 & [0.88,\ 1.03] & $-$0.19 \\
\addlinespace
\multicolumn{4}{l}{\textit{Party-level predictors}} \\
\quad Cabinet party                & 0.62 & [0.54,\ 0.71] & $-$1.56 \\
\quad Populism index (+1 SD)       & 1.45 & [1.28,\ 1.64] & $+$1.79 \\
\quad Economic right (+1 SD)       & 1.15 & [1.03,\ 1.28] & $+$0.60 \\
\quad Right-wing populism int.     & 1.01 & [0.89,\ 1.15] & $+$0.06 \\
\addlinespace
\multicolumn{4}{l}{\textit{Country-level predictors}} \\
\quad Liberal democracy (+1 SD)    & 0.84 & [0.72,\ 0.98] & $-$0.65 \\
\quad $\Delta$Libdem within-country (+1 SD) & 0.969 & [0.954,\ 0.985] & $-$0.12 \\
\bottomrule
\end{tabular}
\end{table}

\paragraph{Note on the Italy--Greece metric divergence.}
Country rankings in the main text use raw start-to-end change in the LOWESS-smoothed series, which need not coincide with the modeled time slopes in Figure~\ref{fig:extdata_country}. Italy is the most divergent case: its modeled slope is marginally the lowest in the sample (posterior mean slightly negative, interval spanning zero), yet its raw LOWESS change is a clear rise ($+$1.8~pp)---the linear slope is more sensitive than the endpoint comparison to within-period fluctuation. Greece shows the opposite: its modeled slope is marginally positive (also not credibly different from zero), while its descriptive LOWESS change is the sample's steepest decline ($-$4.7~pp). Because the two metrics summarize different features of the temporal trajectory, they can diverge for countries whose series is non-monotonic or highly variable within the period; both are valid.
\clearpage

\subsection*{SI Section 2: Cross-instrument validation using Perspective API toxicity scores}

To assess whether our party-level findings depend on the GPT-4.1 instrument, we re-estimated the model using an entirely independent measure: mean Perspective API toxicity per party--country--month cell. Perspective provides continuous 0--1 scores from a neural classifier trained on human-annotated online comments---a wholly separate measurement pipeline, and thus a stringent convergent-validity test of the party-level structure.

The scope of this check is deliberately cross-sectional. Perspective has two documented properties that make it unsuitable for the longitudinal and cross-national comparisons at the heart of the primary analysis: temporal non-stationarity, with multiple undisclosed retrainings~\citep{hartmann2026bye,florio2020time}, and uneven cross-language coverage, with systematic issues identified in German~\citep{nogara2025toxic}, and low performance on on morphologically complex (Finnish, Latvian, Turkish) and lower-resource languages~\citep{sap2019risk,dixon2018measuring}. These are among the reasons we did not adopt it as the primary instrument. They do not, however, compromise its use as an independent check on \textit{party-level} associations, which are identified within time and language. The Perspective panel covers 9,892 of the 10,445 main-model cells (94.7\%), across all 26 countries and 234 parties. We model cell-level mean toxicity (clipped to $(10^{-4},\,1-10^{-4})$ for stability) with a beta-regression likelihood, $y \sim \mathrm{Beta}(\mu\phi,(1-\mu)\phi)$, retaining the main model's linear predictor.

\begin{figure}[ht]
\centering
\includegraphics[width=0.85\textwidth]{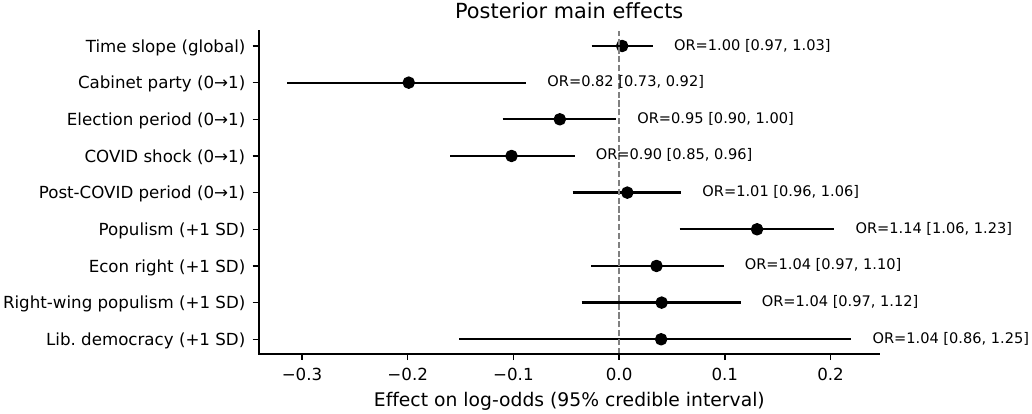}
\caption{\textbf{Perspective API robustness model.} Forest plot of posterior medians and 95\% credible intervals (log-odds scale) for the beta-regression with mean Perspective toxicity as the dependent variable. The party-level populism and cabinet effects replicate the primary model; the time trend and liberal democracy coefficient do not, consistent with Perspective's known temporal and cross-language limitations (see text).}
\label{fig:si_persp}
\end{figure}

The party-level findings replicate on this independent instrument (Supplementary Figure~\ref{fig:si_persp}; Supplementary Table~\ref{tab:si_persp}): populism remains a positive, credible predictor (OR~=~1.14, 95\%~CI: 1.06--1.23), and cabinet membership is associated with lower toxicity (OR~=~0.82, 95\%~CI: 0.73--0.92). The two predictors that do \textit{not} replicate are precisely those that Perspective's known limitations would compromise. The global time trend is null (OR~=~1.00, 95\%~CI: 0.97--1.03), as expected given that Perspective's retraining-induced non-stationarity corrupts longitudinal comparison and its English-skewed coverage attenuates scores in the non-English countries that contribute much of the primary trend. The liberal democracy coefficient reverses (OR~=~1.04 vs.\ 0.84 in the primary model), consistent with the same coverage gap: weaker-democracy countries disproportionately use lower-resource languages, where Perspective systematically under-scores toxicity, masking the cross-sectional association. Both non-replications thus fall on the temporal and cross-national dimensions Perspective cannot measure reliably, while the within-time, within-language party-level structure---the object of this check---holds across both instruments.

\begin{table}[ht]
\centering
\caption{\textbf{Perspective API robustness model.} Beta-regression with mean Perspective toxicity per cell as the dependent variable. $n = 9{,}892$ cells, 26 countries, 234 parties. Odds ratios are exponentiated logit-link coefficients; credible intervals from the posterior.}
\label{tab:si_persp}
\begin{tabular}{lrrr}
\toprule
\textbf{Predictor} & \textbf{OR} & \textbf{95\% CI} & \textbf{$\Delta p$ (pp)} \\
\midrule
\multicolumn{4}{l}{\textit{Time and event effects}} \\
\quad Time slope (per SD)         & 1.00 & [0.97,\ 1.03] & $+$0.02 \\
\quad COVID-19 shock               & 0.90 & [0.85,\ 0.96] & $-$0.53 \\
\quad Post-COVID period            & 1.01 & [0.96,\ 1.06] & $+$0.04 \\
\quad Election period              & 0.95 & [0.90,\ 1.00] & $-$0.30 \\
\addlinespace
\multicolumn{4}{l}{\textit{Party-level predictors}} \\
\quad Cabinet party                & 0.82 & [0.73,\ 0.92] & $-$0.99 \\
\quad Populism index (+1 SD)       & 1.14 & [1.06,\ 1.23] & $+$0.76 \\
\quad Economic right (+1 SD)       & 1.04 & [0.97,\ 1.10] & $+$0.19 \\
\quad Right-wing populism int.     & 1.04 & [0.97,\ 1.12] & $+$0.23 \\
\addlinespace
\multicolumn{4}{l}{\textit{Country-level predictor}} \\
\quad Liberal democracy (+1 SD)    & 1.04 & [0.86,\ 1.25] & $+$0.23 \\
\bottomrule
\end{tabular}
\footnotesize{\\ Note: Populism and cabinet effects replicate the primary GPT-4.1 model directionally and credibly. The null time trend and reversed liberal democracy coefficient reflect Perspective's temporal non-stationarity and English-skewed language coverage, respectively (see text), and are expected under those known limitations rather than evidence against the primary findings.}
\end{table}

\subsection*{SI Section 3: Threshold sensitivity --- any incivility ($\geq 1$)}

The primary analysis uses the ``severe incivility'' threshold (score $\geq 2$), capturing content classified as very or extremely uncivil on the 0--3 GPT-4.1 scale. We chose this threshold because it isolates the content most consequential for democratic discourse and yields a well-separated binary outcome, but it captures only the tail of the distribution. Here we show that the substantive findings do not depend on it.

\begin{figure}[ht]
\centering
\includegraphics[width=0.90\textwidth]{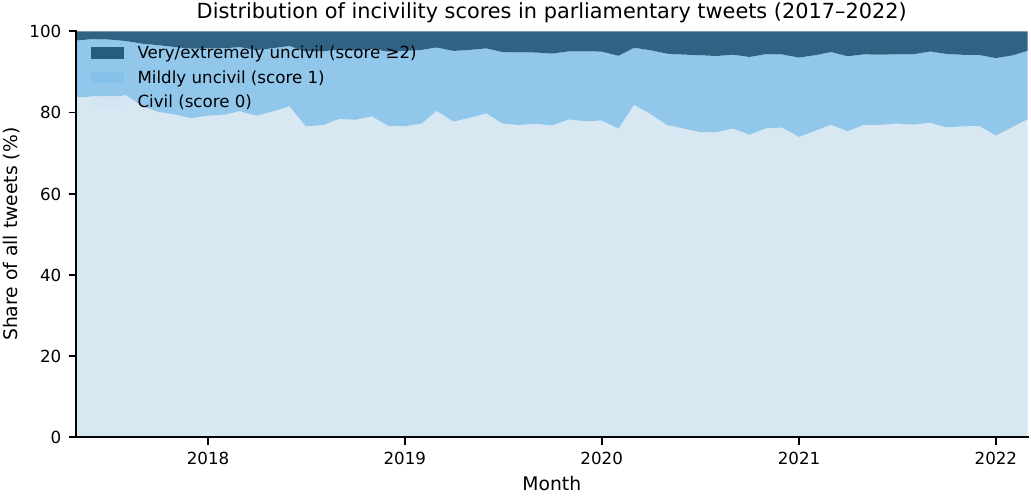}
\caption{\textbf{Incivility score distribution over time.} Stacked area chart showing the monthly share of parliamentary tweets at each incivility level (civil = 0, mildly uncivil = 1, very/extremely uncivil = $\geq 2$), 2017--2022. Approximately 22.4\% of tweets receive any incivility score; the severe threshold used in the primary analysis captures the darkest band (5.1\% on average).}
\label{fig:si_stacked}
\end{figure}

\begin{figure}[ht]
\centering
\includegraphics[width=0.90\textwidth]{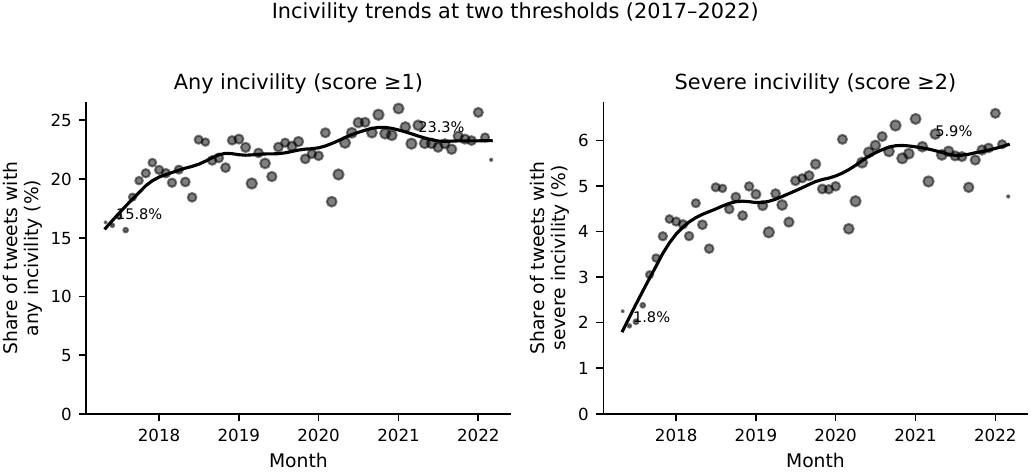}
\caption{\textbf{Incivility trends at two thresholds.} Left: share scoring $\geq 1$ (``any incivility''), LOWESS endpoints $\approx$15.8\% and $\approx$23.3\%. Right: share scoring $\geq 2$ (``severe incivility''), LOWESS endpoints $\approx$1.8\% and $\approx$5.9\%. These literal smoother endpoints are more edge-sensitive than the 2017-average-vs-2021/22-average comparisons used in the main text ($\approx$2.7\% to $\approx$5.8\% for severe; $\approx$17.9\% to $\approx$23.5\% for any incivility); both summaries describe the same rise. Both panels use LOWESS smoothers (bandwidth = 0.25). The parallel trends indicate that the documented rise is not an artifact of threshold choice.}
\label{fig:si_thresholds}
\end{figure}

Figure~\ref{fig:si_stacked} places the threshold in context. Across the study period, approximately 22.4\% of parliamentary tweets receive any incivility score ($\geq 1$)---17.3\% scoring 1 (mildly uncivil) and 5.1\% scoring $\geq 2$ (severe)---leaving 77.6\% classified as civil. Both the mild and severe bands trend upward over time, though the relative increase is more pronounced for severe incivility (Figure~\ref{fig:si_thresholds}).

Re-estimating the hierarchical beta-binomial model with the $\geq 1$ threshold as the dependent variable (base rate 22.4\%, rising from $\approx$17.9\% in 2017 to $\approx$23.5\% by 2021--2022) reproduces the main structural findings (Table~\ref{tab:si_any}). Populism (OR~=~1.25, 95\%~CI: 1.12--1.39) and the cabinet/opposition gap (OR~=~0.49, 95\%~CI: 0.43--0.55) both hold, the latter somewhat larger at this threshold. The time slope remains positive and credible (OR~=~1.06, 95\%~CI: 1.01--1.10) but is attenuated relative to the severe-incivility estimate, as is the COVID shock---consistent with the descriptive finding that the upward trend and the pandemic dip are both sharpest for the most hostile content. The one effect that does not replicate is liberal democracy, which is null at $\geq 1$ (OR~=~1.01, 95\%~CI: 0.90--1.14): institutional constraints appear to bear specifically on the most extreme rhetorical violations, not on ordinary political sharpness.

\begin{table}[ht]
\centering
\caption{\textbf{Threshold sensitivity: any incivility ($\geq 1$).} Main model re-estimated with the share of tweets scoring $\geq 1$ per party--country--month cell as the dependent variable, alongside the primary $\geq 2$ model. Identical hierarchical beta-binomial specification throughout.}
\label{tab:si_any}
\small
\begin{tabular}{lcccc}
\toprule
 & \multicolumn{2}{c}{\textbf{Severe ($\geq 2$)}} & \multicolumn{2}{c}{\textbf{Any ($\geq 1$)}} \\
\cmidrule(lr){2-3}\cmidrule(lr){4-5}
\textbf{Predictor} & \textbf{OR} & \textbf{95\% CI} & \textbf{OR} & \textbf{95\% CI} \\
\midrule
\textit{Time trend} \\
\quad Time slope (global, per SD)        & 1.15 & [1.08,\ 1.21] & 1.06 & [1.01,\ 1.10] \\
\addlinespace
\textit{Within-observation predictors} \\
\quad Cabinet party ($0\to1$)            & 0.62 & [0.53,\ 0.71] & 0.49 & [0.43,\ 0.55] \\
\quad Election period ($0\to1$)          & 0.94 & [0.87,\ 1.01] & 0.87 & [0.82,\ 0.92] \\
\quad COVID shock ($0\to1$)              & 0.91 & [0.82,\ 1.01] & 0.87 & [0.81,\ 0.93] \\
\quad Post-COVID period ($0\to1$)        & 0.91 & [0.83,\ 0.99] & 0.97 & [0.91,\ 1.03] \\
\addlinespace
\textit{Party-level predictors} \\
\quad Populism index ($+1$~SD)           & 1.46 & [1.29,\ 1.65] & 1.25 & [1.12,\ 1.39] \\
\quad Econ.\ left--right ($+1$~SD)       & 1.15 & [1.03,\ 1.28] & 1.10 & [1.00,\ 1.20] \\
\addlinespace
\textit{Country-level predictor} \\
\quad Liberal democracy ($+1$~SD)        & 0.84 & [0.72,\ 0.98] & 1.01 & [0.90,\ 1.14] \\
\bottomrule
\end{tabular}
\footnotesize{\\ Note: OR = odds ratio; 95\% CI = posterior credible interval. Populism and cabinet effects replicate closely across thresholds, with a somewhat larger cabinet/opposition gap at $\geq 1$. The time slope and COVID shock remain credible at $\geq 1$ but attenuated. The liberal democracy association does not replicate at $\geq 1$, consistent with institutional constraints bearing most clearly on the most extreme incivility.}
\end{table}

\begin{figure}[ht]
\centering
\includegraphics[width=0.90\textwidth]{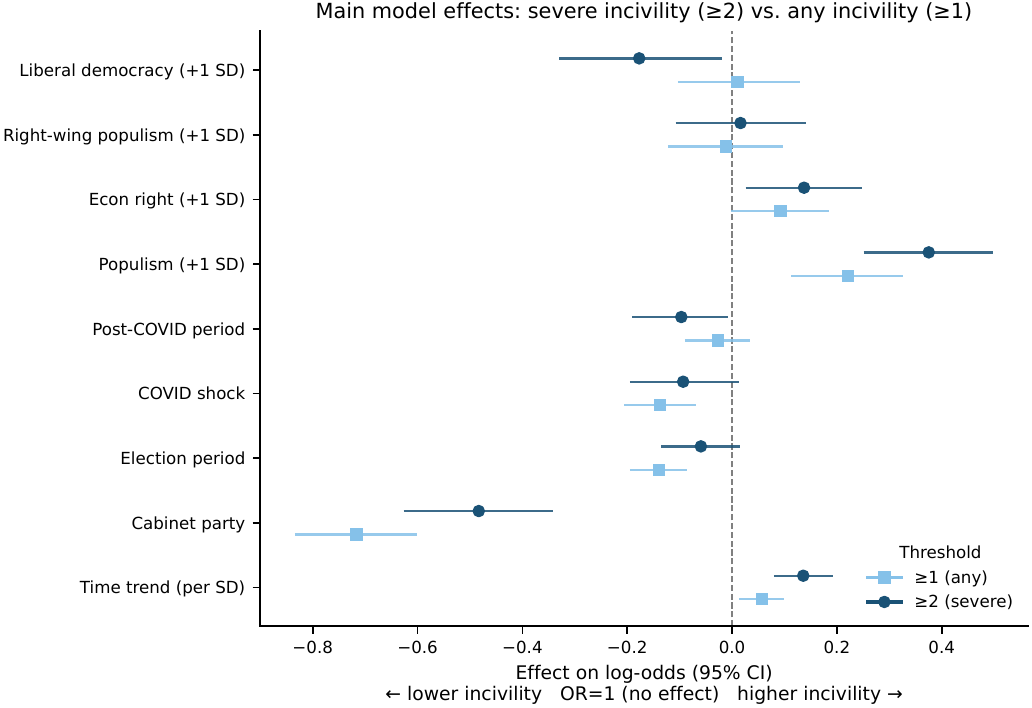}
\caption{\textbf{Model effects at two incivility thresholds.} Side-by-side forest plot comparing posterior medians and 95\% credible intervals (log-odds scale) for all main model predictors at the severe ($\geq 2$, circles) and any ($\geq 1$, squares) thresholds. The direction and approximate magnitude of all main effects replicate across thresholds. The exception is liberal democracy, credibly negative at $\geq 2$ but null at $\geq 1$---consistent with institutional constraints operating most clearly on the extreme end of the distribution.}
\label{fig:si_threshold_comp}
\end{figure}
\clearpage

\subsection*{SI Section 4: LLM classifier validation}

\subsubsection*{Incivility classification prompt and scoring rubric}
The text below is the system prompt sent verbatim with every GPT-4.1 API call (temperature 0); line breaks within paragraphs are typesetting only and not present in the original. Tweets were submitted in a separate user message, batched 50 per call as plain-text \texttt{id}/\texttt{text} pairs; the model received no party, politician, or country metadata.

\begin{quote}
\ttfamily\small
\noindent
You are coding political incivility in tweets by politicians. Code only the
author's communicative posture, not whether you agree with the claim or
whether the claim is true.
Assign one score per tweet: 0, 1, 2, or 3.
SCALE
0 = Civil\\
No incivility. The tweet may criticize policies, decisions, events, parties,
or performance, but it does not use insults, contemptuous language,
profanity, derogatory labels, or delegitimizing attacks.\\
Examples: policy criticism, factual claims, announcements, disagreement,
condolences, praise, campaign updates.
1 = Mildly uncivil\\
Dismissive, sarcastic, mocking, condescending, or mildly contemptuous
language, but no direct strong attack. Also code 1 for profanity used only
as emphasis rather than as a targeted insult.\\
Examples: "predictably," "what a surprise," "they should be ashamed," "this
is crap," "another brilliant idea from..."
2 = Very uncivil\\
Direct attack on a person, party, institution, or group. Includes explicit
insults, derogatory labels, accusations of lying, corruption, betrayal,
malice, incompetence, criminality, hypocrisy, or deliberate deception;
targeted profanity; and delegitimizing opponents as enemies, traitors,
criminals, racists, extremists, etc.\\
Examples: "liar," "corrupt," "traitor," "racist," "crooks," "goons,"
"idiots," "they are destroying the country," "this party is a criminal
organization."
3 = Extremely uncivil\\
Severe norm violation: dehumanization, explicit hate speech, threats,
intimidation, calls for violence, endorsement/celebration of harm, wishes
of death/destruction, or categorical exclusion of groups from
society/politics.\\
Examples: "parasites," "vermin," "they should be killed," "wipe them out,"
"nuke them," "they have no place in our society," explicit
racist/religious hate, calls for expulsion of an entire group.
DECISION RULES
1. Code the highest level of incivility present. One strongly uncivil
phrase is enough.\\
2. Strong criticism is not automatically uncivil. "The policy failed" = 0.
"The minister is incompetent" = 2.\\
3. Profanity always raises the score: non-targeted profanity = at least 1;
targeted profanity/abuse = at least 2.\\
4. Sarcasm/mockery without direct attack = usually 1. Sarcasm used to make
a direct attack = 2.\\
5. Attacks on institutions, parties, media, courts, governments, or social
groups count the same as attacks on individuals.\\
6. Derogatory language toward ethnic, religious, gender, sexual,
immigrant, or other social groups is at least 2; hate speech,
dehumanization, or calls for exclusion/violence are 3.\\
7. If the tweet quotes or reports uncivil language by others in order to
condemn it, code the author's posture, not the quoted content. Usually
this is 0, but note that graphic quoted abuse was present.\\
8. If the tweet reports threats or violence without endorsing them, code
the reporting posture, usually 0.\\
9. If the tweet is unclear, only a link, only a hashtag with no clear
incivility, or impossible to evaluate, code 0.\\
10. When genuinely unsure between two scores, choose the lower score.
Use ids exactly as shown (strings). Output ONLY JSON:\\
\{"labels":[["id",score],...]\} (every id once)
\end{quote}

\paragraph{Classification procedure and reproducibility.} Tweets were classified via the OpenAI Batch API in requests of 50 tweets each, run on 23 May 2026 using the \texttt{gpt-4.1} alias, which resolves to the single released snapshot \texttt{gpt-4.1-2025-04-14} (see Methods); the model is therefore stable across the run. Responses were parsed as a JSON \texttt{labels} array of \texttt{[id, score]} pairs; the small number of requests returning malformed output were excluded rather than imputed, and do not enter the analysis sample. The complete classification pipeline---prompt, batching, parsing, and submission code---is available in the repository (see Code and data availability), allowing the full labelling to be reproduced.

\subsubsection*{Incivility classifier}

\paragraph{Sample design.}
The classifier was validated against a stratified random sample of 1,200 tweets (1,195 successfully coded; five were unavailable or deleted between selection and coding). The sample was boundary-weighted for the most stringent possible test---450 tweets at score 1 and 450 at score 2, bracketing the 1/2 decision boundary used in the main analysis, plus 150 each at scores 0 and 3---concentrating validation power where it matters most for the dependent variable. Within each stratum, tweets were stratified by language family (Germanic, Romance, Hellenic/Turkic, and other) for proportional coverage across the corpus's 20-plus languages.

\paragraph{Coding team.}
Four trained coders annotated the sample independently, blind to the model's scores and to one another. The team was assembled for native-language coverage:
\begin{itemize}
  \item \textbf{Coder 1} (Swedish native speaker): all 205 Germanic-language tweets (German, Dutch, Swedish, Danish, Norwegian) in the original language, plus English and other items; 668 total.
  \item \textbf{Coder 2} (Portuguese native speaker): all Portuguese tweets natively, plus English, Spanish, Italian, and French items without machine translation; 592 total.
  \item \textbf{Coder 3} (Turkish native speaker): all 87 Turkish tweets in the original language---specialist validation for the corpus's most linguistically distant language.
  \item \textbf{Coder 4} (Greek native speaker): all 46 Greek tweets natively.
\end{itemize}
\noindent In total, 899 items (75\%) were coded by a coder working in the tweet's original language without machine translation. The remaining 301 (25\%)---chiefly Polish, Finnish, and Latvian, for which no native coder was available---used DeepL assistance. Each coder received a standardized codebook defining the four-point scale, with corpus examples for every score level and language family and explicit rules for difficult cases (irony, religious rhetoric, indirect attacks); coders had unlimited time and could flag items for discussion. Coders 1 and 2 double-coded a shared 197-item subset spanning all four language families, yielding inter-coder agreement of $\kappa = 0.77$ (quadratic-weighted)---substantial agreement for an inherently subjective construct, and evidence that the codebook produced consistent interpretations.

\paragraph{Agreement statistics.}
Because the sample is boundary-weighted---oversampling the ambiguous 1/2 boundary to stress-test the instrument---raw agreement understates performance on the actual corpus, where most tweets are unambiguously civil. We therefore report both raw agreement and agreement reweighted to the true corpus distribution (83\% score-0, 12\% score-1, 4.5\% score-2, 0.5\% score-3); the reweighted figure estimates expected LLM--human agreement in the full corpus. Across all four coders ($n = 1{,}195$), raw quadratic-weighted $\kappa$ is 0.74, rising to 0.80 reweighted---at or above the two primary coders' own inter-coder agreement ($\kappa = 0.77$), so the classifier achieves agreement comparable to the inter-coder benchmark. Per-coder reweighted $\kappa$ is 0.85 (Coder 1), 0.77 (Coder 2), 0.80 (Coder 3, Turkish), and 0.63 (Coder 4, Greek); three of four, including the Turkish specialist, match or exceed the primary coders, with the Greek sample the exception. At the binary $\geq 2$ threshold, the classifier is conservative---precision 0.84, recall 0.65 ($F_1 = 0.73$)---and 97\% of classifications fall within one ordinal category of the human label. Table~\ref{tab:si_val_kappa} gives reweighted $\kappa$ by language family.

\begin{table}[ht]
\centering
\caption{\textbf{Incivility classifier validation by language family.} Population-reweighted quadratic-weighted $\kappa$ between GPT-4.1 and human coders, by language family. Population reweighting adjusts for the boundary-weighted validation design. Coder 1 (Germanic native speaker) covered all language families; Coder 2 (Portuguese native speaker) covered English, Romance, and Other (not Germanic). Turkish and Greek rows reflect Coders 3 and 4 (native speakers) coding directly in the original language. $n$ = items coded per stratum.}
\label{tab:si_val_kappa}
\small
\begin{tabular}{lccc}
\toprule
\textbf{Language family} & \textbf{$\kappa$ (reweighted)} & \textbf{$\kappa$ (raw)} & \textbf{$n$} \\
\midrule
English                  & 0.76 & 0.72 & 349 \\
Germanic (DE, NL, SV, DA, NO) & 0.83 & 0.76 & 205 \\
Romance (FR, ES, PT, IT) & 0.84 & 0.68 & 228 \\
Turkish (native coding)  & 0.80 & 0.89 & 86 \\
Greek (native coding)    & 0.63 & 0.70 & 46 \\
Other (excl.\ Turkish, Greek) & 0.82 & 0.72 & 281 \\
\midrule
\textbf{Overall}         & \textbf{0.80} & \textbf{0.74} & \textbf{1{,}195} \\
\bottomrule
\end{tabular}
\footnotesize{\\ Note: Population reweighting uses the true corpus distribution: 83\% score-0, 12\% score-1, 4.5\% score-2, 0.5\% score-3. Inter-coder $\kappa$ (Coders 1~\&~2 on 197 shared items): 0.77. 75\% of all validated items were coded in the tweet's original language without machine translation. Agreement is lower for Greek, the one language coded by a specialist whose results diverge from the rest of the sample; this is an acknowledged limitation of the validation.}
\end{table}

\paragraph{Sensitivity to DeepL assistance.}
Since 25\% of items were DeepL-assisted rather than natively coded, we checked whether that subset shows worse agreement. It does not: reweighted $\kappa$ is 0.81 for the 298 DeepL-assisted items versus 0.80 for the 897 natively-coded items (raw: 0.70 and 0.75). Agreement within individual DeepL-assisted languages is more variable (Polish raw $\kappa = 0.61$, $n=91$; Finnish $0.51$, $n=26$), but these single-language estimates are too small to separate genuine differences from noise. The aggregate comparison, adequately powered, shows no sign that machine-assisted coding degraded validation quality.

\paragraph{Country-level calibration check.}
A distinct concern is whether cross-national differences in classifier agreement could masquerade as real differences in incivility---if a country's tweets are disproportionately in a lower-agreement language, its estimated level could reflect measurement error rather than behavior. We assigned each country a language-agreement proxy (the reweighted $\kappa$ for its dominant tweeting language family; the four ``Other''-dominant countries---Poland, Finland, Latvia, Slovenia---take that subset's $\kappa = 0.73$) and correlated it with each country's estimated main-model intercept. The proxy is not credibly associated with the intercept ($r = -0.17$, $p = 0.40$ against the liberal-democracy-adjusted residual; $r = -0.27$, $p = 0.17$ unadjusted; Spearman $\rho = 0.03$, $p = 0.89$); if anything the sign runs opposite to a measurement artifact, with lower agreement associated with \textit{lower} estimated incivility. This is necessarily approximate---one agreement value per country, and 26 countries offer limited power---but it provides no evidence that the cross-national pattern is an artifact of language-varying calibration.

\paragraph{Party-family and populism calibration check.}
Most consequentially, we tested whether disagreement varies with party family or populism in a way that could bias the main estimates---for instance, if the classifier over-flagged populist or radical-right speech. Using the binary $\geq 2$ classification for all 1,195 items (overall disagreement 14.6\%), disagreement is somewhat higher for nationalist/radical-right tweets (19.9\%) than for social-democratic or conservative parties (12.1--13.5\%; Table~\ref{tab:si_partisan_cal}). A logistic regression of disagreement on language-family dummies plus a radical-right indicator gives OR~=~1.48 (bootstrap 95\%~CI: 0.90--2.27, $p = 0.11$), not credible at conventional thresholds; the continuous populism score gives a modest, credible per-SD OR of 1.22 (95\%~CI: 1.03--1.43).

Crucially, this excess disagreement runs in the reassuring direction. For radical-right tweets it is driven by \textit{under}-classification---the model misses severe content humans flag in 13.5\% of cases, versus 6.4\% where it flags content humans do not---and across the populism distribution the model under-classifies at roughly twice the rate it over-classifies. The consequence is that the classifier if anything \textit{underestimates} the incivility of radical-right and high-populism parties, so the populism and radical-right effects in the main model are conservative lower bounds, not artifacts of over-flagging.

\begin{table}[ht]
\centering
\caption{\textbf{Binary LLM--human disagreement by party family.} Over-classifies'' = LLM scores $\geq 2$, human $< 2$. Under-classifies'' = human $\geq 2$, LLM $< 2$. Only families with $n \geq 30$ are shown. Logistic regression controlling for language family yields a non-significant radical-right dummy (OR~=~1.48, bootstrap 95\%~CI: 0.90--2.27), and a significant but modest populism effect (per-SD OR~=~1.22, 95\%~CI: 1.03--1.43). The dominant error direction is under-classification in all groups.}
\label{tab:si_partisan_cal}
\small
\begin{tabular}{lcccc}
\toprule
\textbf{Party family} & \textbf{$n$} & \textbf{Disagree} & \textbf{LLM over-classifies} & \textbf{LLM under-classifies} \\
\midrule
Nationalist/Radical right & 141 & 19.9\% & 6.4\% & 13.5\% \\
Radical left              &  89 & 16.9\% & 4.5\% & 12.4\% \\
Liberal                   & 176 & 18.2\% & 4.0\% & 14.2\% \\
Conservative              & 304 & 13.5\% & 4.9\% &  8.6\% \\
Social democrat           & 305 & 12.1\% & 2.6\% &  9.5\% \\
Christian democrat        &  46 & 13.0\% & 2.2\% & 10.9\% \\
\midrule
\textbf{Overall}          & 1{,}195 & 14.6\% & 3.9\% & 10.6\% \\
\bottomrule
\end{tabular}
\footnotesize{\\ Note: Green/Ecologist ($n=34$), Agrarian ($n=14$), Single issue ($n=19$) omitted due to small sample.}
\end{table}

\subsubsection*{Topic classifier}
The GPT-4.1-mini topic classifier was validated against a 600-tweet sample balanced across the nine categories and language families. Two coders annotated overlapping portions so every item received at least one label, with 204 items double-coded before either saw the other's annotations or the model's output. The initial twelve categories were merged into the final nine (Economy = Economic + Welfare; Cultural Politics = Identity + Nationalism; Law \& Order = Crime + Security) to resolve boundaries that proved conceptually fuzzy in coding; the other six were unchanged. The codebook is available in the data repository.

Following standard practice, the gold standard restricts LLM--human agreement to the 173 double-coded items on which the two coders agreed ($84.8\%$ of the overlap), excluding ambiguous boundary cases. On this gold standard the classifier achieves $\kappa = 0.80$ (82.7\% exact); inter-coder agreement on the full overlap is $\kappa = 0.85$. Because gold-standard restriction excludes the hardest items by construction, we also scored the classifier against all 204 double-coded items, including the 31 disagreement cases, against each coder's label individually: $\kappa = 0.78$ (Coder 1) and $0.71$ (Coder 2)---modestly lower, as expected. Language-specific statistics are in Table~\ref{tab:si_topic_kappa}.

\begin{table}[ht]
\centering
\caption{\textbf{Topic classifier validation by language family.} Cohen's $\kappa$ between GPT-4.1-mini topic assignments (merged nine-category scheme) and the human-coder gold standard (items where both coders independently agreed), by language family. No separate Hellenic- or Turkic-specialist topic coding was conducted; all non-English, non-Germanic, non-Romance items are grouped under ``Other.''}
\label{tab:si_topic_kappa}
\small
\begin{tabular}{lcc}
\toprule
\textbf{Language family} & \textbf{$\kappa$} & \textbf{$n$} \\
\midrule
English    & 0.78 & 76 \\
Germanic   & 0.77 & 35 \\
Romance    & 0.85 & 45 \\
Other (incl.\ TR, EL, PL) & 0.78 & 17 \\
\midrule
\textbf{Overall} & \textbf{0.80} & \textbf{173} \\
\bottomrule
\end{tabular}
\footnotesize{\\ Note: $n$ and $\kappa$ are restricted to the gold-standard subset where both human coders independently agreed (173 of the 204 double-annotated items within the 600-item validation sample). Inter-coder $\kappa$ on the full double-coded overlap (n=204, not gold-standard restricted): 0.85.}
\end{table}

\clearpage

\subsection*{SI Section 5: Country-exclusion robustness for the liberal democracy effect}

Turkey occupies a markedly lower position on V-Dem's liberal democracy index than the rest of the sample throughout the study period (mean $\approx 0.12$ vs.\ 0.50--0.89), raising the concern that the cross-sectional liberal democracy effect could be an artifact of this single extreme case. We address this by refitting the main model with each of the four highest-leverage countries excluded in turn: Turkey and Poland (the two lowest on liberal democracy) and Denmark (the highest), plus Greece (the lowest-agreement language in classifier validation, a related concern about measurement quality; Supplementary Information Section~4). Because a coefficient on a cross-sectional country-level predictor is most sensitive to the extremes of that predictor, these four exclusions are the informative ones; a full 26-way sweep adds little beyond them.

The effect is stable across all four refits (Supplementary Table~\ref{tab:si_loco}). Excluding Turkey does not weaken it---if anything the point estimate strengthens slightly (OR~=~0.80 vs.\ 0.84)---the opposite of what an artifact of Turkey's extreme position would produce. Excluding Denmark or Greece leaves the estimate essentially unchanged. Excluding Poland produces the only credible interval that includes 1.0 (OR~=~0.87, 95\%~CI: 0.74--1.02), but the point estimate barely moves from the full-sample value: it is the interval that widens, not the effect that vanishes---the expected consequence of removing a high-leverage observation from a country-level predictor estimated across only 26 countries. The direction and approximate magnitude of the association hold throughout.

\begin{table}[ht]
\centering
\caption{\textbf{Leave-one-country-out refits of the liberal democracy coefficient.} Each row excludes the named country and refits the full main model (specification otherwise identical). Zero divergences in all four refits.}
\label{tab:si_loco}
\small
\begin{tabular}{lccc}
\toprule
\textbf{Model} & \textbf{OR} & \textbf{95\% CI} & \textbf{$n$ (obs / countries)} \\
\midrule
Full sample (main model)   & 0.84 & [0.72, 0.98] & 10{,}445 / 26 \\
Excluding Turkey           & 0.80 & [0.69, 0.94] & 10{,}072 / 25 \\
Excluding Poland           & 0.87 & [0.74, 1.02] & 10{,}080 / 25 \\
Excluding Denmark          & 0.84 & [0.71, 0.98] & 9{,}903 / 25 \\
Excluding Greece           & 0.84 & [0.72, 0.98] & 10{,}080 / 25 \\
\bottomrule
\end{tabular}
\footnotesize{\\ Note: Liberal democracy ($+1$~SD) coefficient from the main beta-binomial model, refit on each restricted sample. Countries selected by leverage on this coefficient (the two lowest and the highest on liberal democracy) and by classifier validation quality (Greece, the lowest-agreement language; Supplementary Information Section~4).}
\end{table}

\clearpage

\subsection*{SI Section 6: Open-source classifier robustness}

Recent work highlights a replicability challenge for LLM-based measurement: because proprietary model weights, versions, and inference parameters can change or be discontinued, results may be hard to reproduce exactly~\citep{barrie2026replication}. We address this in part through data sharing and through the human-coded validation and cross-instrument checks above (SI Sections~2 and~4), which show the substantive patterns are not artefacts of the specific classifier or prompt. To support replicability further, we trained and publicly release an open-source XLM-RoBERTa-large classifier~\citep{conneau2020unsupervised}---a fully transparent alternative that researchers can re-run, extend, or audit---and refit the main model on its predictions.

\paragraph{Training and evaluation.}
The classifier was fine-tuned in two stages. In Stage~1, XLM-RoBERTa-large was fine-tuned on a stratified 10\% subsample of the GPT-4.1-labelled corpus, using focal loss to handle class imbalance. In Stage~2, the resulting checkpoint was further fine-tuned on all 1,695 human-coded tweets (the 1,195-item primary validation sample plus 500 additional coded tweets), evaluated by 5-fold cross-validation. Across held-out folds, the open classifier achieves F1-macro~=~0.796 (F1-severe~=~0.740; F1-not-severe~=~0.851) against human labels; GPT-4.1 achieves F1-macro~=~0.835 on the same 1,695-tweet human benchmark---a valid comparison because GPT-4.1 was never fine-tuned on these tweets. Applied at $p > 0.5$ to the full 13.8-million-tweet corpus, the open classifier yields a panel-level severe rate of 8.0\% (vs.\ 5.1\% for GPT-4.1), reflecting its lower precision at the decision boundary.

\paragraph{Panel construction.}
We rebuilt the party--country--month panel by substituting the open classifier's binary predictions for the GPT-4.1 score-$\geq$2 threshold, holding all other construction choices fixed. The resulting panel is identical in structure to the main panel: 10,445 observations, 243 parties, 26 countries, 2017--2022.

\paragraph{Results.}
Three of the four key effects replicate with credible intervals excluding 1 (Supplementary Table~\ref{tab:si_opencls}): the time trend (OR~=~1.09, 95\%~CI: 1.04--1.15), cabinet suppression (OR~=~0.61, 95\%~CI: 0.54--0.70), and populism (OR~=~1.31, 95\%~CI: 1.18--1.45) are all directionally consistent and credible under the open classifier. The liberal democracy coefficient is directionally consistent (OR~=~0.89) but its interval spans 1 (0.75--1.07). This attenuation is the expected consequence of the open model's lower precision: approximately uniform classification noise inflates within-cell variance without shifting between-cell means, and so dilutes cross-sectional signals---which rest on between-country contrasts---more than the time-trend and party-level signals, which are estimated from within-unit variation over far more observations. The effect-size ordering (populism $>$ time $>$ liberal democracy) is preserved across both instruments.

\begin{table}[ht]
\centering
\caption{\textbf{Main model refit with open-source XLM-RoBERTa-large classifier.} Columns 2--3 reproduce the GPT-4.1 main model; columns 4--5 report the same model with the panel built from the open classifier's predictions. Specification otherwise identical.}
\label{tab:si_opencls}
\small
\begin{tabular}{lcccc}
\toprule
 & \multicolumn{2}{c}{\textbf{GPT-4.1 (main model)}} & \multicolumn{2}{c}{\textbf{XLM-RoBERTa-large}} \\
\cmidrule(lr){2-3}\cmidrule(lr){4-5}
\textbf{Predictor} & \textbf{OR} & \textbf{95\% CI} & \textbf{OR} & \textbf{95\% CI} \\
\midrule
Time slope (global, per SD) & 1.15 & [1.08,\ 1.21] & 1.09 & [1.04,\ 1.15] \\
Cabinet party               & 0.62 & [0.53,\ 0.71] & 0.61 & [0.54,\ 0.70] \\
Election period             & 0.94 & [0.87,\ 1.01] & 0.97 & [0.91,\ 1.03] \\
COVID shock                 & 0.91 & [0.82,\ 1.01] & 0.87 & [0.80,\ 0.95] \\
Post-COVID period           & 0.91 & [0.83,\ 0.99] & 0.92 & [0.85,\ 0.99] \\
Populism index (+1 SD)      & 1.46 & [1.29,\ 1.65] & 1.31 & [1.17,\ 1.45] \\
Econ left--right (+1 SD)    & 1.15 & [1.03,\ 1.28] & 1.10 & [1.00,\ 1.20] \\
RW-populism interaction (+1 SD) & 1.02 & [0.90,\ 1.15] & 1.02 & [0.92,\ 1.14] \\
Liberal democracy (+1 SD)   & 0.84 & [0.72,\ 0.98] & 0.89 & [0.75,\ 1.07] \\
\bottomrule
\end{tabular}
\footnotesize{\\ Note: $n = 10{,}445$ cells, 243 parties, 26 countries in both models. Severe rate 5.1\% (GPT-4.1) vs.\ 8.0\% (XLM-RoBERTa). XLM-RoBERTa F1-macro 0.796 estimated by 5-fold cross-validation; GPT-4.1 F1-macro 0.835 on the full held-out validation set (the two figures use different evaluation protocols and are not directly comparable).}
\end{table}
\clearpage
\bibliographystyle{unsrtnat}
\bibliography{references}

\end{document}